%% file: main.tex
\documentclass{article} 
\usepackage{iclr2027_conference,times}

\input{math_commands.tex}

\usepackage{hyperref}
\usepackage{url}

\usepackage{longtable}
\usepackage{booktabs}
\usepackage{wrapfig}
\usepackage{wrapfig}
\usepackage{caption}
\usepackage{booktabs,tabularx,array}
\usepackage{float}
\usepackage{listings}
\usepackage{fvextra} 
\usepackage{subcaption}
\usepackage{enumitem}
\usepackage{comment}
\usepackage[most]{tcolorbox}

\usepackage{amsmath}
\usepackage{amssymb}

\usepackage{algorithm}
\usepackage{algpseudocode}

\usepackage{graphicx}

\definecolor{myyellow}{RGB}{255,255,180}
\definecolor{myred}{RGB}{220,0,0}
\definecolor{revision}{RGB}{0, 0, 255}

\input{progress}

\title{Persona Learning from Feedback for Personalized AI Assistants}

\author{Yiran Shen$^{1}$, Yu Xia$^1$, Liuyi Yao$^2$, Prithviraj Ammanabrolu$^1$
 \\
$^1$UC San Diego \quad  $^2$Alibaba Group\\
}

\iclrfinalcopy 
\begin{document}

\maketitle

\input{contents/0_abstract}
\input{contents/1_intro}
\input{contents/2_related}

\input{contents/3_background}
\input{contents/4_method}
\input{contents/5_exp}
\input{contents/6_conclusion}

\newpage
\subsection*{AI use statement}

In this work, we used generative AI tools to assist with language editing, including improving grammar, readability, and clarity of the manuscript. We have not used generative AI tools for generating research ideas, conducting analyses, interpreting results, or formulating scientific conclusions, and other disclosure-related uses are not applicable to this work. Additionally, generative AI tools were used to refine the presentation and wording of selected passages without altering their substantive meaning. All AI-assisted text was carefully reviewed and edited by the authors to ensure that it accurately reflects our original ideas, findings, and interpretations. We take full responsibility for the final content of this work, including all text, claims, conclusions, and any material produced with the aid of generative AI.

\subsection*{Ethics statement}

This work uses existing research datasets, model-generated feedback, simulated tutoring interactions, and a small human evaluation. The human study was approved by the relevant Institutional Review Board (IRB) before data collection, and all participants provided informed consent. Our persona categories are derived from response-style preferences rather than demographic or sensitive attributes. The tutoring experiments use simulated students, so their satisfaction measures should not be interpreted as evidence of human learning outcomes. More broadly, personalization systems may incorrectly infer user preferences or reduce them to coarse categories, and we do not intend PLF for high-stakes profiling or decision making.

\subsection*{Reproducibility statement}

We provide the algorithm, model and training configurations, data processing and split procedures, routing and decoding settings, evaluation protocols, prompts, and statistical procedures needed to reproduce our experiments. Appendices~\ref{app:plf_impl}--\ref{app:tutor-prompts} give implementation, data, training, routing, and prompt details, while Appendices~\ref{app:preference-results}--\ref{app:kl-ablation} provide the full evaluation results and ablation protocols. We also report model versions, random seeds, hardware settings, and dataset revisions where applicable.



\bibliography{iclr2027_ref}
\bibliographystyle{iclr2027_conference}

\appendix
\input{contents/7_appendix}

\end{document}

%% file: math_commands.tex
\usepackage{amsmath,amsfonts,bm}

\def\eqref#1{equation~\ref{#1}}

\def\1{\bm{1}}

\DeclareMathAlphabet{\mathsfit}{\encodingdefault}{\sfdefault}{m}{sl}
\SetMathAlphabet{\mathsfit}{bold}{\encodingdefault}{\sfdefault}{bx}{n}



%% file: progress.tex
\usepackage{colortbl}                 
\usepackage{soul}                     
\usepackage{tcolorbox}
\tcbuselibrary{skins,breakable}

\newif\ifprogress\progresstrue        

\definecolor{pgDoneTx}{HTML}{0B6E33}\definecolor{pgDoneBg}{HTML}{CFEFD8}
\definecolor{pgPendTx}{HTML}{9A2313}\definecolor{pgPendBg}{HTML}{FAD8D2}
\definecolor{pgPartTx}{HTML}{8A5A00}\definecolor{pgPartBg}{HTML}{FBE9C6}
\definecolor{pgInfoTx}{HTML}{1D4ED8}\definecolor{pgInfoBg}{HTML}{DCE6FF}

\ifprogress

  \newtcolorbox{donecard}[1]{breakable,enhanced,boxrule=0.4pt,arc=2pt,
    left=6pt,right=6pt,top=3pt,bottom=3pt,
    colframe=pgDoneTx,colback=pgDoneBg!45,coltitle=white,colbacktitle=pgDoneTx,
    fonttitle=\bfseries\footnotesize,fontupper=\footnotesize,
    title={ACHIEVED \textbf{\textperiodcentered}\ #1}}
  \newtcolorbox{pendcard}[1]{breakable,enhanced,boxrule=0.4pt,arc=2pt,
    left=6pt,right=6pt,top=3pt,bottom=3pt,
    colframe=pgPendTx,colback=pgPendBg!45,coltitle=white,colbacktitle=pgPendTx,
    fonttitle=\bfseries\footnotesize,fontupper=\footnotesize,
    title={PENDING \textbf{\textperiodcentered}\ #1}}
  \newtcolorbox{partcard}[1]{breakable,enhanced,boxrule=0.4pt,arc=2pt,
    left=6pt,right=6pt,top=3pt,bottom=3pt,
    colframe=pgPartTx,colback=pgPartBg!45,coltitle=white,colbacktitle=pgPartTx,
    fonttitle=\bfseries\footnotesize,fontupper=\footnotesize,
    title={PARTIAL \textbf{\textperiodcentered}\ #1}}
  \newtcolorbox{infocard}[1]{breakable,enhanced,boxrule=0.4pt,arc=2pt,
    left=6pt,right=6pt,top=3pt,bottom=3pt,
    colframe=pgInfoTx,colback=pgInfoBg!50,coltitle=white,colbacktitle=pgInfoTx,
    fonttitle=\bfseries\footnotesize,fontupper=\footnotesize,
    title={CODEBASE EVIDENCE (not yet in paper) \textbf{\textperiodcentered}\ #1}}
  \newtcolorbox{dashcard}[1]{breakable,enhanced,boxrule=0.8pt,arc=3pt,
    left=7pt,right=7pt,top=5pt,bottom=5pt,
    colframe=black!80,colback=black!3,coltitle=white,colbacktitle=black!80,
    fonttitle=\bfseries\small,fontupper=\footnotesize,
    title={#1}}

\else
\fi

%% file: contents/0_abstract.tex


\begin{abstract}
Personalizing AI assistants requires adapting their behavior to users' preferences and task contexts.
Existing work elicits assistant personas through persona prompting or finetuning on persona-conditioned generations.
However, persona descriptions alone do not identify what to change or preserve in a particular response, or which behaviors suit the current request.
We introduce \emph{Persona Learning from Feedback (PLF)}, a framework that learns reusable, preference-grounded assistant personas as independent LoRA adapters on a shared frozen backbone.
During training, a frozen self-teacher conditions on persona specifications and response-specific feedback to provide token-level supervision on each adapter's own generated prefixes.
A feedback gate focuses corrective updates on responses judged to need improvement.
At deployment, a supervised router uses the request or dialogue to select a single adapter or a fixed composition, without user-specific finetuning.
We evaluate PLF on single-turn personalization with model-generated critiques and multi-turn tutoring with simulated students' reactions.
Across both settings, PLF improves model-judged preference fit over fixed-persona policies and strong baselines including persona prompting and persona distillation, demonstrating how feedback-trained behaviors can be reused for adaptive assistant personalization without degrading general capabilities.

\end{abstract}

%% file: contents/1_intro.tex

\section{Introduction}
\label{sec:introduction}

Personal AI agents are designed to support users across ongoing tasks and goals, using context accumulated through interaction.
Effective assistance in these settings depends not only on what users want to accomplish, but also on how they prefer to be helped~\citep{kirk2024the,sorensen2024roadmap}.
The same user may want a concise, actionable recommendation when making an urgent decision, but a detailed explanation when exploring an unfamiliar topic.
We study this behavioral dimension of personalization: adapting how an assistant helps to the preferences and task requirements expressed in the current request or dialogue.

We represent reusable patterns of assistance through \emph{assistant personas}.
Here, a persona specifies preferred response behaviors and behaviors to avoid, such as how much detail to provide or whether to explain a solution directly or guide the user toward it.
We ground persona specifications in user preferences, allowing behaviors to be reused across users and tasks and combined when a request calls for multiple preferences~\citep{jang2023personalized}.
For example, a user may prefer an answer that is both warm and concise.

Existing work elicits assistant personas through role or style prompting~\citep{xu2023expertprompting,kong-etal-2024-better} and finetuning on persona-conditioned generations~\citep{maiya2026open}, with recent approaches also selecting personas according to task utility~\citep{hu2026expert}.
Personalization methods further incorporate retrieved user history~\citep{salemi2024lamp}, learn from heterogeneous user feedback~\citep{li2025personalized}, or infer response preferences from interaction history~\citep{spohn2026apm}.
These approaches can specify or infer desired behaviors, but a persona description alone does not identify which parts of a particular response should be revised or preserved.
For example, when a user requests concise troubleshooting advice, one draft may bury the solution in unnecessary background, while another may omit an essential step. Both require different corrections despite sharing the same persona description.



Natural-language feedback can supply these response-specific corrections.
Prior work uses critiques to refine outputs~\citep{madaan2023selfrefine,akyurek2023rl4f} or converts textual feedback into training signals through self-distillation~\citep{hubotter2026reinforcement}.
For personalization, however, the desired correction is itself preference-dependent: shortening an answer may improve it for one request while making another less useful.
This creates two complementary problems: \emph{learning} reusable personas from preference-dependent, response-specific feedback, and \emph{selecting} the appropriate persona or combination from the evidence available before each response.


\begin{figure*}[t]
    \centering
    \includegraphics[width=0.9\linewidth]{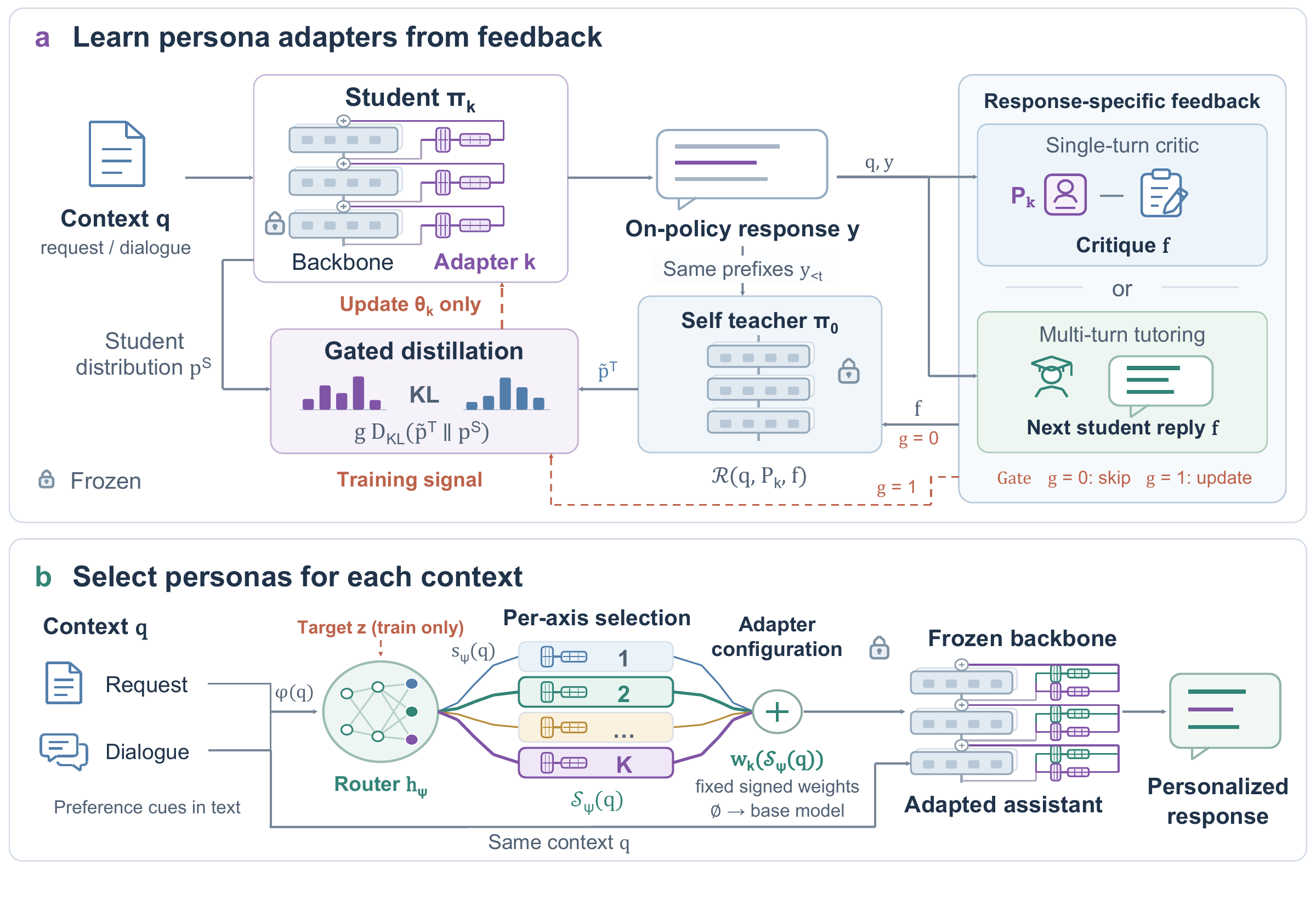}
    \vspace{-1em}
    \caption{
\textbf{Persona Learning from Feedback (PLF).}
\textbf{(a)} A frozen teacher uses critiques or conversational feedback to supervise persona adapters on student-generated prefixes.
A feedback gate excludes satisfactory responses from updates.
\textbf{(b)} A supervised router selects personas using the request or dialogue, activating a single adapter or a fixed signed composition. 
}
    \label{fig:method}
\end{figure*}

We introduce \emph{Persona Learning from Feedback (PLF)}, a framework that separates feedback-based persona learning from request-specific persona selection (Figure~\ref{fig:method}).
Each persona is represented by an independent LoRA adapter on a shared frozen backbone~\citep{hu2022lora}, making its learned parameter updates separately selectable and composable.
During training, an adapter generates its own responses and receives feedback evaluated against its intended persona or the corresponding user preference.
A feedback gate focuses corrective updates on responses judged to need improvement.
A frozen self-teacher conditions on the persona specification and feedback to provide token-level supervision on the student's own generated prefixes~\citep{agarwal2024policy,zhao2026self}.
The adapter thus learns to express its persona without an additional persona prompt at inference.
After persona learning, the adapters are frozen and a supervised router identifies the desired personas from preference evidence expressed in the request or dialogue.
A single selected persona activates its adapter and paired selections use a signed weight-space composition with coefficients chosen on validation data.
This allows the same adapters to serve different users and requests without per-user finetuning or separate adapters for every persona combination.

We evaluate PLF in single-turn personalization and multi-turn tutoring.
The single-turn setting uses PRISM requests with assigned preferences grounded in recurring patterns among training users, and model-generated critiques provide response-specific supervision~\citep{kirk2024the}.
The tutoring setting uses simulated students' subsequent replies as feedback on preceding tutor responses, starting from openings in the Education Dialogue Dataset~\citep{shani2024multiturn}.
We compare PLF with fixed-persona policies and persona prompting, alongside two training alternatives: persona distillation and feedback imitation. We also examine reuse under combined preferences and preference changes during dialogue.
Ablations study the effects of feedback gating and the distillation objective.
We measure model-judged preference fit in both settings. In tutoring, we additionally report simulated student satisfaction and independently judged factual errors.
Across both settings, PLF improves average model-judged preference fit over strong baselines.
Adapter composition improves preference fit over individual heads for selected preference pairs, and the router can reuse the same heads as preferences change during dialogue.
Our contributions are as follows:
\begin{itemize}
    \item We formulate assistant personalization as learning reusable, preference-grounded personas from response-specific feedback and selecting them using observable context, without per-user finetuning.
    \item We introduce PLF, which learns independent persona adapters through feedback-gated on-policy self-distillation and reuses them through supervised routing and signed composition.
    \item We demonstrate improved preference fit over fixed-persona policies, persona prompting, persona distillation, and feedback imitation in both single-turn personalization and multi-turn tutoring, and analyze combined preferences, preference changes during dialogue, and training ablations.
\end{itemize}





%% file: contents/2_related.tex
\section{Related Work}
\label{sec:related_work}

\paragraph{Assistant personas and user personalization.}
Persona and role prompting can steer LLM behavior through instructions about role, expertise, style, or interaction strategy, though effects vary across models and tasks~\citep{xu2023expertprompting,salewski2023incontext,kong-etal-2024-better,zheng-etal-2024-helpful}. Such behaviors can also be internalized through training. Open Character Training learns persistent character traits from preference data and synthetic dialogues~\citep{maiya2026open}, while \citet{hu2026expert} identify a trade-off between persona expression and task accuracy and introduce PRISM to selectively activate a distilled persona adapter. Complementary work personalizes behavior using user information. LaMP retrieves from user history~\citep{salemi2024lamp}, Personalized-RLHF learns from heterogeneous user feedback~\citep{li2025personalized}, and APM evaluates preference inference from interaction history while reducing reliance on stereotypical associations~\citep{spohn2026apm}. Together, these works separate eliciting a behavioral tendency from deciding when it should be used.


\paragraph{Learning from feedback and on-policy distillation.}
Natural-language feedback can improve model behavior through inference-time critique and revision~\citep{madaan2023selfrefine,akyurek2023rl4f}, or through training on feedback-informed revisions and optimization signals derived from critiques~\citep{chen2024learning,wang2026textgrad}. On-policy distillation provides teacher supervision on student-generated trajectories, reducing mismatch between training data and the policy's own outputs~\citep{agarwal2024policy}. Methods condition the teacher on privileged information unavailable to the deployed student~\citep{zhao2026self}, including textual feedback in SDPO and RLTF-SD~\citep{hubotter2026reinforcement,song2026expanding}. Related work applies feedback learning to personalization by distilling information from subsequent user messages~\citep{buening2026aligning} or using natural-language feedback to improve personalized question answering~\citep{salemi2026learning}.


\paragraph{Modular personalization and routing.}
Parameter-efficient and modular adaptation offers an alternative to representing all preferences in a single policy. Personalized Soups trains models for different preference dimensions and combines them through post-hoc parameter merging~\citep{jang2023personalized}, while mixture-of-experts methods select or combine LoRA modules~\citep{wu2024mixture}. For personalized generation, PROPER uses group-level LoRA adaptations with routing over shared and user-specific preferences~\citep{zhang-etal-2025-proper}, and CARD combines cluster-level LoRA adaptation with lightweight user-specific personalization~\citep{song2026card}. These approaches motivate modular representations for heterogeneous users and preferences, while leaving open when modularization is preferable to a shared conditional policy.

%% file: contents/4_method.tex
\section{Method}
\label{sec:method}

\emph{Persona Learning from Feedback (PLF)} separates learning how to realize reusable personas in individual responses from selecting the appropriate personas for each request or interaction.
Persona specifications define the intended behaviors, response-specific feedback guides their learning, and a supervised router selects personas from the observable request or dialogue at deployment.
The selected set determines a single adapter or a fixed composition of adapters (Figure~\ref{fig:method}).
Algorithm~\ref{alg:plf} summarizes the procedure.

Let $q$ denote the context available before an assistant response, including the current request and any preceding dialogue.
PLF assumes a fixed catalog $\mathcal P=\{P_k\}_{k=1}^{K}$, where each $P_k$ specifies a reusable assistance behavior associated with a user preference profile.
Each persona is represented by an independent LoRA adapter~\citep{hu2022lora}, called a \emph{persona head}, with parameters $\theta_k$ on a shared frozen base model $\pi_0$, yielding policy $\pi_k$.
Separate adapters keep the learned parameter updates independently selectable and composable.
All adapters start with zero weight updates, so $\pi_k=\pi_0$ before training, and each learns its assigned behavior across training contexts.
We withhold persona descriptions from the student prompt to train the adapters to express their behaviors without persona prompting.
The descriptions instead guide feedback and teacher supervision.
At deployment, both the router and the response generator receive $q$, without a separately supplied preference profile.

\subsection{Response-Specific Feedback}
\label{sec:plf_feedback}

A persona specification describes the intended behavior but does not identify how a particular response should change.
We obtain this response-specific guidance from feedback evaluated against the persona or its corresponding user preference.
For persona $k$, the current policy samples $y\sim\pi_k(\cdot\mid q)$ and receives natural-language feedback $f$ and an update gate $g\in\{0,1\}$~\citep{agarwal2024policy}.

\textbf{Single-turn critiques.} A critic observes $(P_k,q,y)$ and identifies changes needed to express the intended persona while preserving useful content and satisfying the task.
Its critique supplies $f$.
It may instead report that the response needs no revision.

\textbf{Conversational reactions.} In multi-turn scenario, the adapter interacts with a user simulator conditioned on the learning preference corresponding to $P_k$.
The user's next reply supplies $f$ directly, without a separate textual critique.
The assistant response and user reaction are also used to determine the update gate.
The reaction is unavailable when generating $y$, but becomes part of the dialogue context for the next response.

\textbf{Feedback gating.} We focus corrective supervision on responses that still need improvement, rather than treating satisfactory responses as further revision targets.
We set $g=0$ when the feedback procedure accepts a response without further correction, and $g=1$ otherwise.
Critic outputs with no usable feedback also receive $g=0$.
Both sources yield examples $(q,y,f,g)$ for the same gated learning procedure.
Feedback prompts and judgment criteria are specified in Section~\ref{sec:exp-setup}.

\subsection{Learning Personas from Feedback}
\label{sec:plf_learning}

\textbf{Feedback-conditioned teacher.} To translate response-level feedback into token-level supervision, we use the frozen base model $\pi_0$ as a self-teacher with privileged context.
Its input $\mathcal R(q,P_k,f)$ augments the original context with the persona specification and feedback as hindsight guidance.
The teacher is instructed to answer the original request rather than respond to the feedback.
The complete sampled response is not inserted into $\mathcal R$ as an earlier draft.
For a minibatch indexed by $\mathcal B_k$, containing examples $(q_i,y_i,f_i,g_i)$, we score both models on the student's prefixes so that supervision applies to states visited by the current policy:
\begin{equation}
    p^{\mathrm S}_{i,t}=\pi_k(\cdot\mid q_i,y_{i,<t}),\qquad p^{\mathrm T}_{i,t}=\pi_0\!\left(\cdot\mid\mathcal R(q_i,P_k,f_i),y_{i,<t}\right).
    \label{eq:plf_distributions}
\end{equation}
The feedback thus informs next-token targets without requiring a separately generated revised answer~\citep{zhao2026self,hubotter2026reinforcement}.

\textbf{Gated distillation.} To reduce target storage, we retain and renormalize the teacher's top-$M$ token probabilities to obtain $\widetilde p^{\mathrm T}_{i,t}$, while keeping the student's full-vocabulary normalization.
The adapter minimizes the gated forward-KL objective:
\begin{equation}
    \mathcal L_{\mathrm{persona}}(\theta_k)=\frac{1}{N_k}\sum_{i\in\mathcal B_k}g_i\sum_{t=1}^{|y_i|}D_{\mathrm{KL}}\!\left(\widetilde p^{\mathrm T}_{i,t}\,\middle\|\,p^{\mathrm S}_{i,t}\right),\qquad N_k=\sum_{i\in\mathcal B_k}|y_i|.
    \label{eq:plf_distillation}
\end{equation}
Here $|y_i|$ counts valid response tokens, excluding padding, and $N_k$ includes these tokens even when $g_i=0$.
Examples with $g_i=0$ contribute no distillation loss.
A batch with no positive gates is skipped.
Responses, feedback, and teacher targets are held fixed during the update, and gradients reach only $\theta_k$.
Feedback collection and distillation alternate in rollout cycles: each cycle collects fresh responses from the current adapters and may perform multiple minibatch updates before the next collection step.
Implementation details appear in Appendices~\ref{app:plf_impl} and~\ref{app:exp-details}.

\subsection{Preference-Based Routing and Deployment}
\label{sec:plf_routing}

To match reusable behaviors to request-specific preferences, we freeze the learned adapters and train a router using evidence in the observable context.

\textbf{Routing supervision.} Routing examples are constructed under known preference profiles, which determine their target personas through the predefined preference-to-persona mapping.
The router learns to infer these targets from the resulting request or dialogue, without observing the profile itself.
Preferences can concern different aspects of assistance, allowing a request to combine behaviors such as a warm tone and a concise explanation.
We therefore organize persona indices into $J$ disjoint pairs $\{\mathcal I_a\}_{a=1}^{J}$, each containing two contrasting alternatives on a preference axis, so $K=2J$.
The routing dataset $\mathcal D_{\mathrm{route}}=\{(q_i,\mathbf z_i)\}_{i=1}^{n}$ contains nonzero targets $\mathbf z_i\in\{0,1\}^{K}$ satisfying $\sum_{k\in\mathcal I_a}z_{i,k}\leq1$ for each axis.

\textbf{Router training and selection.} A lightweight classifier $h_\psi$ maps representations $\varphi(q)$ from the frozen base model, with adapters disabled, to $K$ logits.
We train it to predict the persona targets:
\begin{equation}
    \mathcal L_{\mathrm{route}}(\psi)=\frac{1}{n}\sum_{i=1}^{n}\ell_{\mathrm{CE}}\!\left(h_\psi(\varphi(q_i)),\mathbf z_i\right).
    \label{eq:plf_router_loss}
\end{equation}
Here $\ell_{\mathrm{CE}}$ is mean binary cross-entropy over personas for multi-label routing or categorical cross-entropy for exclusively single-label routing.
Only $\psi$ is trained.
Applying sigmoid for multi-label routing or softmax for single-label routing to these logits gives the scores $\mathbf s_\psi(q)$.
To avoid selecting contrasting behaviors together, we retain at most one persona per axis, choosing its highest-scoring alternative when the score exceeds a fixed threshold:
\begin{equation}
    \begin{aligned}
    \widehat k_a(q) &= \operatorname*{arg\,max}_{k\in\mathcal I_a}s_{\psi,k}(q),\\
    \mathcal S_\psi(q) &= \left\{\widehat k_a(q):a=1,\ldots,J,\,s_{\psi,\widehat k_a(q)}(q)>\tau\right\}.
    \end{aligned}
    \label{eq:plf_selection}
\end{equation}

\textbf{Adapter configuration.} The selected personas determine fixed adapter coefficients, while router scores are used only for selection.
At an adapted layer $\ell$, let $W_\ell^0$ be the base weight matrix and $\Delta W_{\ell,k}$ persona $k$'s LoRA update.
Generation uses
\begin{equation}
    W_\ell(q)=W_\ell^0+\sum_{k=1}^{K}w_k\!\left(\mathcal S_\psi(q)\right)\Delta W_{\ell,k}.
    \label{eq:plf_composition}
\end{equation}
A single selection activates only its corresponding adapter with coefficient one, while an empty selection uses the base model as a fallback.
To support combined preferences without training a separate adapter for every pair, we compose the learned updates in weight space.
Given a selected pair $\mathcal S=\{k_a,k_b\}$ from distinct axes, let $\bar k_c$ denote the opposite persona to $k_c$ on axis $c$.
For each axis, we start at the midpoint of its two adapter updates and add $\lambda_c$ times the contrast from the opposite to the selected persona.
Averaging the two resulting axis-specific updates gives the coefficients
\begin{equation}
    w_{k_c}(\mathcal S)=\frac14+\frac{\lambda_c}{2},\qquad w_{\bar k_c}(\mathcal S)=\frac14-\frac{\lambda_c}{2},\qquad c\in\{a,b\},
    \label{eq:plf_pair_coefficients}
\end{equation}
with all other coefficients zero.
The coefficients sum to one, and $\lambda_c>1/2$ gives the opposite persona a negative coefficient.
The strengths $\lambda_c$ are fixed for each selected pair $\mathcal S$ before evaluation, with tuned values chosen on validation data rather than predicted by the router.
The signed sum is represented exactly as one adapter through the LoRA factor concatenation described in Appendix~\ref{app:plf_impl}.
The adapter configuration remains fixed during a response and is recomputed from the updated context before each subsequent response.
Section~\ref{sec:exp-setup} specifies the persona catalog, routing-data construction, classifier and selection configurations, and composition-strength settings.

\begin{algorithm}[t]
\small
\caption{Persona Learning from Feedback (PLF)}
\label{alg:plf}
\begin{algorithmic}[1]
\Require Frozen base $\pi_0$, personas $\{P_k\}$, context and feedback sources, and routing data $\mathcal D_{\mathrm{route}}$.
\Statex \textbf{Persona learning}
\State Initialize adapters with zero weight updates.
\For{each training cycle and each persona $k$}
    \State Collect a rollout set $\mathcal E_k=\{(q_i,y_i,f_i,g_i)\}$ on-policy from $\pi_k$.
    \For{each minibatch indexed by $\mathcal B_k$ drawn from $\mathcal E_k$}
        \State $N_k\gets\sum_{i\in\mathcal B_k}|y_i|$.
        \If{$\sum_{i\in\mathcal B_k}g_i>0$}
            \State Compute teacher targets $\widetilde p^{\mathrm T}_{i,t}$ on $y_{i,<t}$ for $g_i=1$.
            \State Update only $\theta_k$ using Eq.~\ref{eq:plf_distillation}.
        \EndIf
    \EndFor
\EndFor
\Statex \textbf{Router training}
\State Freeze adapters and compute $\varphi(q_i)$ with $\pi_0$ on $\mathcal D_{\mathrm{route}}$.
\State Train $\psi$ using Eq.~\ref{eq:plf_router_loss}.
\Statex \textbf{Deployment for each context $q$}
\State Score $\mathbf s_\psi(q)$ and select $\mathcal S_\psi(q)$ using Eq.~\ref{eq:plf_selection}.
\State Generate from $q$ with fixed weights $w_k(\mathcal S_\psi(q))$ using Eq.~\ref{eq:plf_composition}.
\end{algorithmic}
\end{algorithm}

%% file: contents/5_exp.tex
\section{Experiments}
\label{sec:experiments}

We evaluate Persona Learning from Feedback (PLF) in single-turn
personalization and multi-turn tutoring. We ask whether natural language
feedback produces useful persona heads, whether those heads support combined
or changing preferences, and how the training design affects their behavior.

\subsection{Experimental Setup}
\label{sec:exp-setup}

\textbf{Single-turn setting: preferences in a request.}
We use opening requests from PRISM~\citep{kirk2024the}. 
Following
\citet{kirk2026prism}, recurring survey preferences among training users
motivate two persona axes: \emph{warm} versus \emph{impersonal} voice, and
\emph{concise} versus \emph{thorough} depth. They capture common differences
in desired tone and detail, while permitting combinations such as warm and
concise. We split requests by user and remove exact and near-duplicate tasks
across splits, yielding 6{,}400 training, 203 validation, and 500 test requests.
For routing and evaluation, we minimally rewrite each request to express one
of four individual personas or four cross-axis pairs~\citep{spohn2026apm}. The assigned preference,
rather than the user's original survey response, is the evaluation target.
During head training, a Qwen3.5-27B critic reads the original request, the
head's answer, and its persona card, then explains what to change and preserve.

\textbf{Multi-turn setting: preferences revealed through interaction.}
\label{sec:setting-tutor}
We use teacher openings from the Education Dialogue
Dataset~\citep{shani2024multiturn}: 4{,}524 training, 200 validation, and 500
test dialogues, with topics held out across splits. All subsequent turns are
generated. Our personas are \emph{direct instruction}, which explains and
demonstrates, and \emph{interactive guidance}, which asks questions and lets
the student reason. A Qwen3.5-27B student
simulator holds a hidden preference and reacts to each tutor turn. The following student's turn
is the feedback, rather than a separate critique. 

\textbf{Training.}
For PLF, we freeze the backbone and train one LoRA head per persona
in both settings, using Qwen3-4B-Instruct-2507~\citep{yang2025qwen3}
for single-turn responses and
Llama-3.1-8B-Instruct~\citep{grattafiori2024llama} for tutoring.
The frozen base, conditioned on the persona card and feedback, teaches the
head on its own sampled response prefixes through forward KL distillation.
A no-change feedback gate excludes satisfactory responses from the loss.
We select single-turn heads on validation and use the final heads from
the tutoring run. After freezing the heads, we train MLP routers
on the backbone's last-token representation: the single-turn router selects
persona indicators from the request, while the tutoring router selects a
head after each student reply. Data, training settings, and prompts details can be found in
Appendices~\ref{app:exp-details}--\ref{app:tutor-prompts}.

\textbf{Evaluation.}
\label{sec:plf-baselines}
In both settings, we compare routed PLF against the frozen base model,
each learned persona head used as a fixed policy, and
router-based \textbf{Persona prompting} \citep{spohn2026apm},
which uses the same router as PLF to select persona
descriptions for the base model's prompt. 
We also compare with \textbf{Persona distillation}, which retains
on-policy distillation~\citep{agarwal2024policy,zhao2026self} for each head with a persona-conditioned teacher
but omits response-specific feedback from the teacher context, and
\textbf{Feedback imitation}~\citep{chen2024learning},
which trains each head by supervised learning on responses generated
by the same persona- and feedback-conditioned teacher as PLF.
For \emph{single-turn composition},
we additionally compare with prompting using both requested personas.
On validation, we also compare three alternative mixing rules: summing
adapter updates, averaging adapter updates, and mixing head logits.
For \emph{multi-turn tutoring}, we include an oracle that
selects the head matching the student's hidden preference, with anticipating
and one-turn-lagged variants for preference switching. Supplementary
tutoring controls include oracle persona prompting, linear routers, and
the base without a persona prompt.
Our primary metric in both settings is \emph{pairwise preference win rate}:
a separate \texttt{gpt-6-luna} judge evaluates which response or complete
dialogue better fits the assigned preference. Each pair is judged in both
presentation orders; disagreements count as ties, and ties receive half
credit. For tutoring, we also report \emph{student satisfaction}, the share
of student replies without flagged objections, and \emph{factual error rate},
the share of tutor turns flagged by a separate fact-checking judge.
Evaluation protocols, decoding settings, and full comparisons appear in
Appendices~\ref{app:exp-statistics} and
\ref{app:tutor-results}.

\subsection{Can Natural Language Feedback Learn Useful Personas?}
\label{sec:rq-feedback}

\begin{figure}[t]
    \centering
    \includegraphics[width=0.9\linewidth]{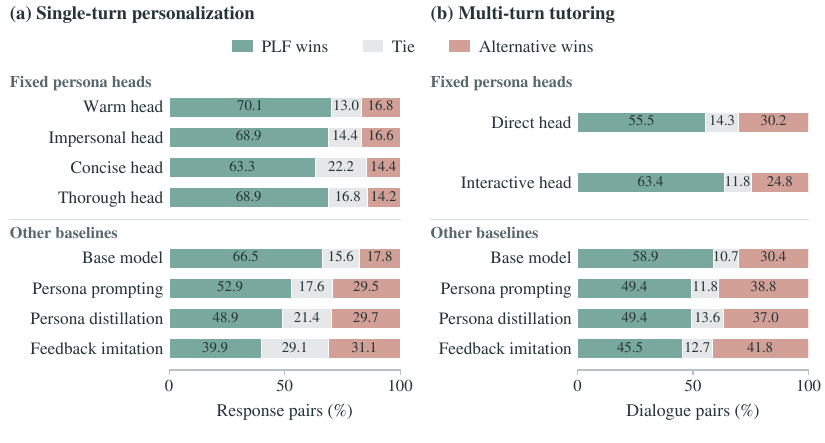}
    \vspace{-0.5em}
    \caption{\textbf{Preference fit of feedback-trained personas.}
    Percentages of PLF wins, ties, and alternative wins on
    (a) 499 single-turn requests and (b) 1{,}000 tutoring dialogue pairs
    from 500 openings evaluated with both student types. Persona prompting
    uses the same router as PLF.}
    \label{fig:persona-learning}
    \vspace{-0.5em}
\end{figure}

\textbf{Learned heads improve on both fixed personas, prompting, and other learning baselines.}
In single-turn personalization, PLF is preferred over every fixed head and
the base model (Figure~\ref{fig:persona-learning}a). No one persona serves
the heterogeneous test preferences as well as selecting and combining
behaviors. PLF also achieves a 61.7\% win rate against persona prompting with
the same router, 59.6\% against persona distillation, and 54.4\%
against feedback imitation. 
The gain over persona distillation shows the value of response-specific
feedback beyond persona-conditioned supervision, while the comparison
with feedback imitation favors PLF's on-policy objective over learning
from complete teacher-generated responses.
The prompting comparison holds preference
selection fixed, showing that learned heads add value beyond
identifying a suitable persona description. The smaller advantage over persona prompting than over
fixed heads is consistent with preference matching contributing substantially
to the full system's gain.

\textbf{Student reactions provide useful feedback without an explicit critique.}
\label{sec:rq-interaction}
Tutoring shows a similar advantage over fixed heads, persona
prompting, and other learning baselines (Figure~\ref{fig:persona-learning}b). PLF also improves over persona distillation with a 56.2\% win rate,
while feedback imitation remains competitive. PLF approaches
the oracle on average, although a correctly matched fixed head can still
outperform it
(Appendix~\ref{app:tutor-results}). This identifies routing as a remaining
source of error: useful heads do not ensure that the right one is always
selected. Compared with persona prompting, PLF also improves student
satisfaction and reduces flagged factual errors, suggesting that its
preference advantage is not achieved by sacrificing measured tutoring
quality. 

\subsection{Can Learned Personas Serve Combined or Changing Preferences?}
\label{sec:rq-routing}

\textbf{Single-turn experiment: combine voice and depth.}
We test composition on requests expressing both a voice and a depth
preference. Heads are trained only on individual personas; the router sees
both single and paired labels during training. At inference, we combine
adapter weights using a validation-selected rule that adjusts voice and
depth strength separately (Appendix~\ref{app:composition-rules}).
Table~\ref{tab:preference-mixing} isolates composition on correctly routed
requests.

\begin{table}[t]
    \centering
    \begingroup
    \captionsetup{font=small,skip=4pt}

    \begin{minipage}[t]{0.49\linewidth}
        \vspace{0pt}
        \caption{\textbf{Single-turn composition.}
        Mixture win rates (\%) against fixed heads. 
        }
        \label{tab:preference-mixing}
    \end{minipage}
    \hfill
    \begin{minipage}[t]{0.49\linewidth}
        \vspace{0pt}
        \caption{\textbf{Multi-turn preference switching.}
        PLF win rates (\%) against alternatives. 
        }
        \label{tab:preference-switching}
    \end{minipage}

    \par

\begin{minipage}[t]{0.49\linewidth}
    \vspace{0pt}
    \centering
    \small
    \setlength{\tabcolsep}{2.2pt}
    \renewcommand{\arraystretch}{1.05}
    \begin{tabular}{@{}lcccc@{}}
        \toprule
        Pair & Warm & Impers. & Conc. & Thorough \\
        \midrule
        Warm + concise
& \cellcolor{blue!5}\textbf{89} & 78
& \cellcolor{blue!5}\textbf{72} & 98 \\
Warm + thorough
& \cellcolor{blue!5}\textbf{81} & 97
& 99 & \cellcolor{blue!5}\textbf{47} \\
Impers. + concise
& 68 & \cellcolor{blue!5}\textbf{51}
& \cellcolor{blue!5}\textbf{53} & 82 \\
Impers. + thorough
& 91 & \cellcolor{blue!5}\textbf{66}
& 81 & \cellcolor{blue!5}\textbf{63} \\
        \bottomrule
    \end{tabular}
\end{minipage}
\hfill
\begin{minipage}[t]{0.49\linewidth}
    \vspace{0pt}
    \centering
    \small
    \setlength{\tabcolsep}{2.5pt}
    \renewcommand{\arraystretch}{1.05}
    \begin{tabular}{@{}lcccc@{}}
        \toprule
        Switch & Direct & Inter. & Base & Oracle \\
        \midrule
        Direct $\to$ interactive
            & \cellcolor{blue!5}\textbf{55} & 81
            & 54 & 47 \\
        Interactive $\to$ direct
            & 67 & \cellcolor{blue!5}\textbf{56}
            & 59 & 40 \\
        \midrule
        Mean
            & 61 & 68
            & 56 & 43 \\
        \bottomrule
    \end{tabular}
\end{minipage}

    \par\smallskip

    \begin{minipage}{\linewidth}
        \footnotesize
        Left: 222/247 paired requests routed correctly, with bold marking constituent heads.
        Right: 500 openings per direction, with bold marking the initially matching head.
        Full results appear in Appendix~\ref{app:preference-results}.
    \end{minipage}

    \endgroup
    \vspace{-1em}
\end{table}

\textbf{Composition enables learned personas to serve combined preferences.}
Warm-concise gives the clearest improvement over both constituent heads, combining brevity with encouragement.
Impersonal-thorough improves over both heads, while warm-thorough improves over warm.
The limited gains for impersonal-concise suggest that composition is most effective when behaviors are complementary.
The warm-concise advantage persists when routing errors are included.
On validation, calibrating voice and depth weights outperforms simple summation or averaging, with weaker voice weights selected for thorough pairs.
The results show that independently learned personas can be reused to serve combined preferences without training separate adapters for each combination.

\textbf{Multi-turn experiment: respond to a preference switch.}
For each test opening, the student's preference switches once, in both
directions, at a uniformly sampled turn $j\in\{2,\ldots,7\}$.
The switch is unannounced: its first evidence is the student's reply to turn
$j$, so the router can first respond at $j+1$. The router is frozen and was
trained only on fixed-preference dialogues. We compare it with fixed heads,
the base, and an oracle that knows the new preference in advance and switches
at turn $j$. The judge sees both preferences and the marked switch point.

\begin{figure}[t]
    \centering
    \includegraphics[width=0.9\linewidth]{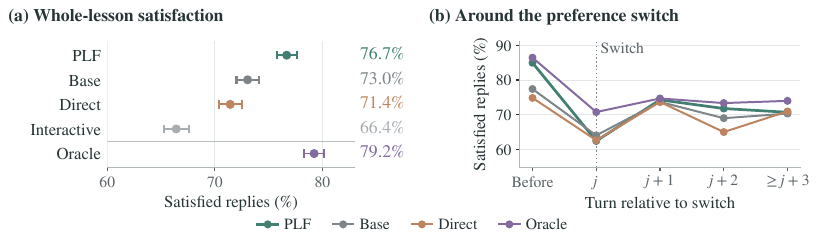}
    \caption{\textbf{Student satisfaction under changing preferences.}
    Share of replies without flagged objections, averaged across switch
    directions: (a) whole lessons, with 95\% bootstrap intervals;
    (b) aligned to the switch at $j$. Only the oracle knows the new
    preference when choosing turn $j$.}
    \label{fig:preference-switching}
    \vspace{-0.5em}
\end{figure}

\textbf{Routing generalizes to changing preferences.}
Although trained only on fixed-preference dialogues, PLF successfully transfers to settings where students change their preferred teaching style during a conversation.
PLF is preferred over both fixed persona heads and the base model in both switch directions (Table~\ref{tab:preference-switching}).
It also achieves the highest whole-lesson satisfaction among non-oracle policies (76.7\%), approaching the oracle (79.2\%) that knows the preference changes in advance (Figure~\ref{fig:preference-switching}a).
Notably, satisfaction rebounds at $j+1$, the first turn at which the router can use student feedback reflecting the new preference (Figure~\ref{fig:preference-switching}b).
These results demonstrate that a router learned from fixed preferences can leverage evolving dialogue context to adapt persona selection without retraining, outperforming static persona assignments under changing user needs.
Further analyses of adaptation dynamics are provided in Appendix~\ref{app:tutor-shift}.

\textbf{Qualitative examples.}
\label{sec:exp-qualitative}
Table~\ref{tab:qualitative-personas} shows how these behaviors appear in responses. For the working-from-home request, the mixture gives a short five-item answer with a friendly tone and is preferred to both constituent heads in both presentation orders. In tutoring, a request for more interactive guidance is followed by a question from the interactive head.

\begin{table}[!t]
    \centering
    \caption{\textbf{Examples of persona composition and adaptation.}
    The warm-head response (129 words in full) is excerpted; emoji are omitted.}
    \label{tab:qualitative-personas}
    \input{figures/qualitative_examples.tex}
\end{table}

\subsection{Ablations: Feedback Gating and KL Direction}
\label{sec:plf-ablations}

We examine whether focusing corrective updates on responses that need improvement helps persona learning, and how the direction of the distillation objective affects preference fit.

\begin{figure}[!t]
    \centering
    \begin{subfigure}[t]{0.45\linewidth}
        \centering
        \includegraphics[width=\linewidth]{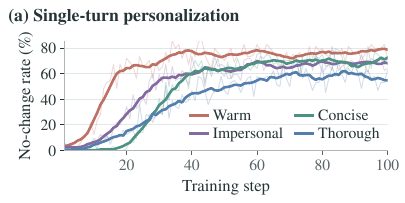}
        \phantomsubcaption
        \label{fig:singleturn-training}
    \end{subfigure}
    \begin{subfigure}[t]{0.45\linewidth}
        \centering
        \includegraphics[width=\linewidth]{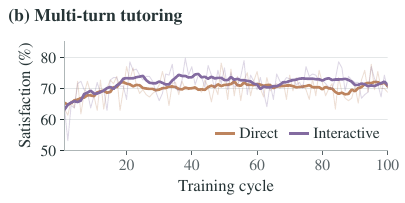}
        \phantomsubcaption
        \label{fig:tutor-training}
    \end{subfigure}
    \vspace{-1.5em}
    \caption{\textbf{Feedback gating during persona learning.}
(a) Critic no-change rate in single-turn personalization.
(b) Satisfaction-judge acceptance in multi-turn tutoring before additional gating checks.
Higher values indicate more responses passing the displayed feedback check.
Thin lines are raw values and thick lines are centered nine-point means.}
    \label{fig:no-change-dynamics}
    \vspace{-1em}
\end{figure}

\textbf{Voice preferences are learned faster than response depth.}
The gated training curves reveal distinct learning dynamics across personas (Figure~\ref{fig:no-change-dynamics}a).
Warm and impersonal adapters reach high critic approval early, whereas concise improves more gradually and thorough remains the most challenging, ending with the lowest approval.
This suggests that adjusting response voice may be easier than controlling response depth: voice preferences can often be expressed through local wording choices, while depth requires balancing the amount of information provided against relevance and completeness.
In multi-turn tutoring, training satisfaction increases rapidly before plateauing (Figure~\ref{fig:no-change-dynamics}b), suggesting that much of the adaptation to the desired teaching behaviors occurs early in training.
These trends reflect training feedback rather than held-out performance; tutoring satisfaction is measured before the additional checks applied by the final gate.

\textbf{Feedback gating improves preference fit through selective updates.}
Feedback gating consistently benefits single-turn personalization, achieving a 61\% win rate over ungated training at matched checkpoints, with improvements across both single-persona and paired-preference requests.
In multi-turn tutoring, gating also improves the interactive persona, achieving a 55\% win rate while encouraging more question-asking and reducing complaints about repetitive responses.
The preference advantage persists on held-out evaluations even as earlier training gains diminish, while the direct persona remains near parity.
Interestingly, gated and ungated models achieve similar critic approval rates despite their differences in preference fit.
This suggests that updating on every response is not necessarily beneficial: selectively focusing learning on responses that require correction can better preserve and refine the desired persona behavior.
Additional analyses are provided in Appendix~\ref{app:gate-ablation}.

\textbf{KL direction has persona-dependent effects.}
At the same step checkpoints, forward KL achieves a 55\% preference win rate against reverse KL over 499 routed single-turn requests.
The advantage is strongest for impersonal-only and impersonal--thorough requests (68\% and 69\%).
Across all requests involving each preference, forward KL also benefits the thorough preference, while warm and concise remain inconclusive.
Under reconstructed teacher contexts, reverse-KL heads have lower entropy, closer to the teacher's, while forward KL retains more probability mass on hedge terms.
Forward-KL impersonal responses use more hedging while reverse-KL thorough responses are longer, but this additional length does not correspond to a preference advantage under the updated judge.
These associations suggest that KL direction can affect persona fit beyond top-token agreement.
Additional analyses are in Appendix~\ref{app:kl-ablation}.

\subsection{General Capability Retention}
\label{sec:capability}

To assess whether improvements in personalization compromise general capabilities, we evaluate the routed systems on IFEval~\citep{zhou2023instruction}, GSM8K~\citep{cobbe2021training}, MMLU-Pro~\citep{wang2024mmlupro}, and MT-Bench~\citep{zheng2023judging}.
Table~\ref{tab:capability-routed} shows that PLF performs comparably to the base model across all four benchmarks in both settings, with no statistically significant degradation.
Unlike feedback imitation, PLF does not exhibit a significant MMLU-Pro drop in single-turn personalization; it also achieves the highest observed MT-Bench score in multi-turn tutoring.
Together with its preference gains, these results show that PLF improves personalization while retaining strong general-purpose performance.

\begin{table}[t!]
\centering
\caption{\textbf{Scores on four standard LLM benchmarks.} We compare the base models with PLF and three alternative baseline methods. Scores are percentages except MT-Bench (1--10).}
\label{tab:capability-routed}
\begingroup
\fontsize{8}{9}\selectfont
\setlength{\tabcolsep}{2pt}
\renewcommand{\arraystretch}{0.95}
\begin{tabular*}{\linewidth}{@{\extracolsep{\fill}}lrrrr@{}}
\toprule
Method & IFEval $\uparrow$ & GSM8K $\uparrow$ & MMLU-Pro $\uparrow$ & MT-Bench $\uparrow$ \\
\midrule
\multicolumn{5}{@{}l}{\textit{Single-turn QA/ Qwen3-4B-Instruct-2507}} \\
Base & 86.5 & 93.2 & 67.2 & 6.87 \\
Persona prompting & 87.2 (+0.7) & 92.8 ($-$0.4) & 67.4 (+0.2) & 6.78 ($-$0.09) \\
Persona distillation & 86.1 ($-$0.4) & 93.4 (+0.2) & 67.4 (+0.2) & 6.68 ($-$0.19) \\
Feedback imitation & 85.6 ($-$0.9) & 92.3 ($-$0.9) & 66.3 ($-$0.9) & 6.69 ($-$0.18) \\
\cellcolor{blue!5}\textbf{PLF} & \cellcolor{blue!5}\textbf{85.4 ({\boldmath$-$}1.1)} & \cellcolor{blue!5}\textbf{93.0 ({\boldmath$-$}0.2)} & \cellcolor{blue!5}\textbf{66.9 ({\boldmath$-$}0.3)} & \cellcolor{blue!5}\textbf{6.79 ({\boldmath$-$}0.08)} \\
\midrule
\multicolumn{5}{@{}l}{\textit{Multi-turn Tutoring / Llama-3.1-8B-Instruct}} \\
Base & 80.2 & 86.6 & 46.9 & 5.66 \\
Persona prompting & 77.6 ($-$2.6) & 85.4 ($-$1.1) & 46.9 ($-$0.0) & 5.72 (+0.07) \\
Persona distillation & 79.1 ($-$1.1) & 86.8 (+0.2) & 47.1 (+0.2) & 5.78 (+0.12) \\
Feedback imitation & 79.3 ($-$0.9) & 86.4 ($-$0.2) & 47.1 (+0.1) & 5.67 (+0.01) \\
\cellcolor{blue!5}\textbf{PLF} & \cellcolor{blue!5}\textbf{78.7 ({\boldmath$-$}1.5)} & \cellcolor{blue!5}\textbf{86.5 ({\boldmath$-$}0.1)} & \cellcolor{blue!5}\textbf{46.4 ({\boldmath$-$}0.5)} & \cellcolor{blue!5}\textbf{5.86 (+0.21)} \\
\bottomrule
\end{tabular*}
\endgroup
\end{table}

%% file: figures/qualitative_examples.tex
\begingroup
\small
\setlength{\tabcolsep}{4pt}
\renewcommand{\arraystretch}{1.15}
\begin{tabularx}{\linewidth}{@{}>{\raggedright\arraybackslash}p{0.22\linewidth}>{\raggedright\arraybackslash}X@{}}
\toprule
\multicolumn{2}{@{}l@{}}{\textbf{(a) Persona composition: warm + concise}} \\
\multicolumn{2}{@{}>{\raggedright\arraybackslash}p{\linewidth}@{}}{\textit{User request:} ``Hi! Could you tell me five benefits of working from home? Keep it friendly and short, please!''} \\
\midrule
\textit{Concise head} & ``1. Save time on commuting. 2. Work in your own rhythm. 3. No office stress. 4. More control over your day. 5. Save money on expenses.'' \\
\addlinespace[3pt]
\textit{Warm head} & ``Absolutely! Here are five cozy, home-based wins-no jargon, just vibes: 1. No commute stress-You skip traffic, parking, and the daily grind. Just chill in your PJs and sip your coffee\ldots'' \\
\addlinespace[3pt]
\textbf{PLF mixture} & ``Sure! Here's a quick list: 1. No commute stress-just save time and energy. 2. Work in your favorite comfy spot. 3. More time for yourself and your family. 4. Flex your schedule to fit your energy. 5. Fewer office distractions-focus, no noise!'' \\
\midrule
\multicolumn{2}{@{}l@{}}{\textbf{(b) Tutoring: direct $\rightarrow$ interactive}} \\
\addlinespace[2pt]
\textit{Student feedback} & ``You're just telling me facts again; \textbf{ask me a real question instead.}'' \\
\addlinespace[3pt]
\textbf{Interactive head} & ``Let's put this into practice: if a car has a mass of 1500 kilograms and is subjected to a force of 3000 Newtons, \textbf{what is its acceleration?}'' \\
\bottomrule
\end{tabularx}
\endgroup

%% file: contents/6_conclusion.tex
\section{Conclusion}
\label{sec:conclusion}

We introduced Persona Learning from Feedback (PLF), which learns reusable persona adapters from response-specific feedback through on-policy self-distillation and selects or combines them using contextual preference evidence.
Across single-turn personalization and multi-turn tutoring, PLF improves model-judged preference fit over fixed persona policies, the base model, persona prompting, and other learning baselines.
The learned personas also support combined and changing preferences. These findings support separating persona learning from contextual selection as an approach to personalization without user-specific parameter updates.

%% file: contents/7_appendix.tex
\newpage

\section{PLF Implementation Details}
\label{app:plf_impl}

\subsection{Truncated Teacher Targets}

Teacher conditioning is defined in Section~\ref{sec:plf_learning}.

For persona $k$'s batch, let $\mathcal V^{(M)}_{i,t}$ contain the indices of the $M$ most probable tokens under $p^{\mathrm T}_{i,t}$.
The stored target is
\begin{equation}
    \widetilde p^{\mathrm T}_{i,t}(v)=\frac{\mathbf 1\{v\in\mathcal V^{(M)}_{i,t}\}p^{\mathrm T}_{i,t}(v)}{\sum_{u\in\mathcal V^{(M)}_{i,t}}p^{\mathrm T}_{i,t}(u)}.
    \label{eq:plf_topm}
\end{equation}
We use $M=256$ to reduce target storage and transfer costs.
The student retains full-vocabulary normalization and is not renormalized over the teacher's retained support.
The teacher target is fixed during each update.
In Eq.~\ref{eq:plf_distillation}, $|y_i|$ excludes padding, and $N_k$ counts valid tokens even when $g_i=0$.
The loss is not renormalized over only the examples selected by the gate.

\subsection{LoRA Parameterization and Exact Composition}

For each adapted weight matrix, persona $k$ contributes
\begin{equation}
    \Delta W_{\ell,k}=\frac{\alpha_{\mathrm L}}{r_{\mathrm L}}B_{\ell,k}A_{\ell,k},
    \label{eq:plf_lora}
\end{equation}
where $A_{\ell,k}\in\mathbb R^{r_{\mathrm L}\times d_{\mathrm{in}}}$ and $B_{\ell,k}\in\mathbb R^{d_{\mathrm{out}}\times r_{\mathrm L}}$.
We initialize $B_{\ell,k}=0$, so each persona policy initially recovers the base model.
Persona adapters have disjoint parameters, and the shared base weights remain frozen.

Let $k_1,\ldots,k_m$ index the adapters with nonzero composition coefficients, and suppress the dependence of the coefficients on $\mathcal S$.
Exact weight-space composition uses
\begin{equation}
    A_\ell^{\mathrm{mix}}=\begin{bmatrix}A_{\ell,k_1}\\\vdots\\A_{\ell,k_m}\end{bmatrix},\qquad B_\ell^{\mathrm{mix}}=\begin{bmatrix}w_{k_1}B_{\ell,k_1}&\cdots&w_{k_m}B_{\ell,k_m}\end{bmatrix}.
    \label{eq:plf_factor_concatenation}
\end{equation}
Keeping the same ratio of LoRA scale to rank gives
\begin{equation}
    \frac{\alpha_{\mathrm L}}{r_{\mathrm L}}B_\ell^{\mathrm{mix}}A_\ell^{\mathrm{mix}}=\sum_{j=1}^{m}w_{k_j}\Delta W_{\ell,k_j}.
    \label{eq:plf_exact_composition}
\end{equation}
The paired implementation on Qwen3-4B uses four adapters with $r_{\mathrm L}=32$ and $\alpha_{\mathrm L}=64$ per adapter.
Concatenation produces a rank-128 parameterization with scale 256, preserving $\alpha_{\mathrm L}/r_{\mathrm L}=2$.
The result is one composite adapter and does not require generating responses from separate policies.

\subsection{Signed Pair Composition}

For the selected pair $\mathcal S=\{k_a,k_b\}$, the coefficients in Eq.~\ref{eq:plf_pair_coefficients} give
\begin{equation}
    \begin{aligned}
    \Delta W_\ell^{\mathrm{pair}}
        =\sum_{c\in\{a,b\}}\Bigl[&\frac14\left(\Delta W_{\ell,k_c}+\Delta W_{\ell,\bar k_c}\right)\\
        &+\frac{\lambda_c}{2}\left(\Delta W_{\ell,k_c}-\Delta W_{\ell,\bar k_c}\right)\Bigr].
    \end{aligned}
    \label{eq:plf_pair_composition}
\end{equation}
This is the mean of the four participating updates plus one weighted contrast per selected axis.
The coefficients sum to one but need not be nonnegative.

\section{Additional Experiment Details}
\label{app:exp-details}

This section describes the data, PLF training configurations, routing procedures, and evaluation protocol used in Section~\ref{sec:experiments}.
Each persona head is an independent LoRA adapter on a frozen backbone.
Both settings use rank 32, LoRA scale $\alpha_{\mathrm L}=64$, no adapter dropout, and adapters on all attention and MLP projections.
Heads sample at temperature 0.7 and are trained with AdamW at learning rate $10^{-5}$ and 3\% warmup.
The frozen backbone supplies feedback-conditioned teacher targets, retaining its top-256 tokens as specified in Appendix~\ref{app:plf_impl}.

\subsection{Single-Turn Data, Training, and Routing}
\label{app:singleturn-details}

\textbf{Data source and preference screening.}
We use the opening requests and survey preferences from PRISM~\citep{kirk2024the}, joining \texttt{conversations.opening\_prompt} and \texttt{survey.system\_string} by \texttt{user\_id}.
Only the first user message is retained from each conversation.
It contains 8{,}011 conversations from 1{,}396 users who entered the conversation phase.

Preferences are manually screened for whether a single response from a model without tools can satisfy them.
We exclude 198 users whose preferences request sourcing or verification and 31 whose preferences primarily concern unavailable capabilities, such as cross-conversation memory, browsing, or real-world actions, and therefore retain 1{,}167 users.

\textbf{Persona catalog.}
Following the use of group-level preferences in \citet{kirk2026prism}, we code the retained preferences for response style, excluding topics, capability requests, and shared requirements such as accuracy, honesty, safety, and respect.
Co-occurring style codes define warm and impersonal voice groups and concise and thorough depth groups.
A user is assigned by the style they emphasize most, or by the first secondary style associated with a group when the primary style has no group.
Users with no grouped style enter a base group, while a separate clarifying-question group is too small to train as a persona.

Group summaries use only training users, with 254 warm, 326 impersonal, 127 concise, and 75 thorough users.
Each summary retains response-style requirements expressed by at least 20\% of the group's training users and omits clauses that cannot be judged from the response without knowing the correct content.
The resulting persona cards each describe one axis and are fixed before the reported training and evaluation.
Their exact wording is given in Appendix~\ref{app:singleturn-personas}.
The original survey preferences and group summaries are not supplied as separate conditioning text to the student, teacher, critic, or router.
The teacher and critic receive the target persona card instead.

\textbf{Splits and request cleaning.}
Splits are stratified by preference group and keep each user's requests together, using the fixed salt \texttt{prism-v2-stratify-v1}.
The 239 users inspected during design are excluded from validation and test.
We remove contentless openings and exact duplicates. Near duplicates have token-set Jaccard similarity at least 0.90, or at least 0.75 with containment at least 0.95 and at least five tokens in the shorter request.
Candidate pairs are blocked using tokens with document frequency at most 100 and grouped into connected components; cross-split matches are excluded without moving requests between splits.
The final splits share no users or exact or near-duplicate requests under these rules (Table~\ref{tab:singleturn-data-splits}).

\begin{table}[htbp]
\centering
\small
\setlength{\tabcolsep}{8pt}
\caption{\textbf{Single-turn data splits.}
User counts refer to the original-request splits after filtering and deduplication.
Rewritten requests are used for router training or evaluation, not for persona-head training.
The final head-training corpus is separately expanded to 6{,}400 original requests from 1{,}242 users.}
\label{tab:singleturn-data-splits}
\begin{tabular}{@{}lrrr@{}}
\toprule
Split & Users & Original requests & Usable rewrites \\
\midrule
Train & 1{,}033 & 5{,}698 & 5{,}677 \\
Validation & 39 & 203 & 203 \\
Test & 89 & 500 & 499 \\
\bottomrule
\end{tabular}
\end{table}

A manual task-quality review removes and replaces 31 requests without a usable task.
The final head-training corpus contains 5{,}671 requests from the original training split and 729 additional openings from users excluded by the preference filter, for 6{,}400 requests from 1{,}242 users.
These additions supply tasks rather than preference labels and are checked against all splits using the same duplicate rules.
Sampling uses seed 20260925 and seed 20260926 for replacements; training uses 100 batches of 64 requests.

\textbf{Preference-embedded requests.}
Following the prompt-rewriting approach of \citet{spohn2026apm}, router training and evaluation use minimally rewritten requests that express an assigned preference through the user's message.
Each original request receives one of eight conditions: four individual personas or four cross-axis pairs.
Conditions are balanced before rewriting using a seeded shuffle with \texttt{crc32('pref-assign|<split>')}.
The assigned condition, rather than the original user's survey response, supplies the router target and the preference judge's reference.
The four condition labels and the rewriting prompt are given in Appendix~\ref{app:singleturn-rewrite-prompts}.

Rewriting uses Qwen3.5-27B with thinking disabled, temperature 1.0, and at most 400 generated tokens.
The seed is \texttt{crc32('pref-rw|<split>|<example\_id>')} plus the attempt index, with up to two retries after a refusal.
After retries, 21 training requests and one test request have no usable rewrite, leaving 5{,}677 training, 203 validation, and 499 test requests.
The test set contains 247 paired-preference requests.

\textbf{Head training.}
The backbone and frozen teacher are Qwen3-4B-Instruct-2507.
Each batch sends the same 64 original requests to all four heads, producing 256 sampled responses with one response per request and head.
The student receives only the request, without a persona card or survey preference.
A Qwen3.5-27B critic receives the target persona card, request, and sampled response as one user message with thinking disabled.
It returns a JSON object containing \texttt{judgement}, \texttt{changes needed}, and \texttt{feedback}.
Only the feedback text conditions the teacher, while the judgment is logged and the change indicator controls the gate.
The critic assesses persona fit and request fulfillment, not factual correctness.

For PLF, the frozen teacher scores the sampled continuation under the persona- and feedback-conditioned context in Appendix~\ref{app:singleturn-training-prompts}, supplying top-256 token targets.
A response contributes zero loss if the critic call fails, no feedback string is parsed, or \texttt{changes needed} is \texttt{No}; its valid tokens remain in the denominator (Appendix~\ref{app:plf_impl}).

Table~\ref{tab:singleturn-training-config} gives the training configuration.
Checkpoints are saved every 20 steps.
We compare steps 40, 60, 80, and 100 on 50 randomly selected validation requests, evaluating each head against its own persona card, and use the step-60 heads for all reported evaluations.

\begin{table}[htbp]
\centering
\small
\renewcommand{\arraystretch}{1.1}
\caption{\textbf{Single-turn persona-head training configuration.}
The backbone remains frozen, and the four heads are independent LoRA adapters.}
\label{tab:singleturn-training-config}
\begin{tabularx}{\linewidth}{@{}lX@{}}
\toprule
Setting & Value \\
\midrule
LoRA & Rank 32, scale 64, dropout 0, all linear layers \\
Training pass & 6{,}400 original requests, 100 batches of 64 \\
Batch sizes & Training 64, training microbatch 4, rollout microbatch 8 \\
Student sampling & Temperature 0.7, top-$p$ 1.0, one sample per head and request \\
Token limits & 1{,}024 prompt tokens, 2{,}560 new response tokens \\
Teacher targets & Top-256 support, full-vocabulary student normalization \\
Optimizer & AdamW, learning rate $10^{-5}$, $\beta=(0.9,0.95)$ \\
Schedule & Constant after 3\% warmup, head learning-rate multiplier 1.0 \\
Regularization & Weight decay 0, gradient-norm clipping 1.0 \\
Critic decoding & Temperature 0, thinking disabled, at most 1{,}000 tokens \\
Training execution & bfloat16, DeepSpeed ZeRO-2, gradient checkpointing, SDPA attention \\
Rollout execution & vLLM, four engines with tensor parallelism 2, colocated with training \\
Hardware & Eight NVIDIA A100-SXM4-80GB GPUs, shared with the critic server \\
Random seed & 7 \\
\bottomrule
\end{tabularx}
\end{table}

\textbf{Router.}
Router features are the frozen backbone's final-layer hidden state after the final normalization, at the last token of the chat-formatted rewritten request.
This token is the newline following the assistant role marker in the generation prefix.
Adapters are disabled when extracting the 2{,}560-dimensional features, which are standardized using training statistics.
The router has layers $2{,}560\rightarrow512\rightarrow512\rightarrow4$, with ReLU activations and dropout 0.1 after each hidden layer.
Its four sigmoid outputs are trained with binary cross-entropy against the assigned persona indicators in the order warm, impersonal, concise, and thorough.
For example, warm plus concise has target $(1,0,1,0)$.

The usable training rewrites are split by user into 5{,}123 fitting requests from 930 users and 554 development requests from 103 users.
The optimizer is AdamW with learning rate $10^{-3}$, weight decay $10^{-2}$, and batch size 128.
Early stopping monitors development binary cross-entropy with patience 15, selecting epoch 4.
For each axis, the higher-scoring persona is selected only when its probability exceeds 0.5.
If neither axis yields a selection, generation uses the base model.
On test, the selected persona set matches the assigned set on 426 of 499 requests, including 222 of 247 paired requests.

One selected persona activates its head alone.
Two selected personas use the signed composition coefficients in Appendix~\ref{app:composition-rules}, without generating separate candidate responses.
For thorough pairs, the voice strength is selected from $\{1,0.5,0.25\}$ using the 50 thorough-pair validation requests.
The checkpoint, router configuration, and composition coefficients are fixed before test evaluation.
At deployment, the router and learned heads receive only the rewritten request, without a separately supplied condition label, survey preference, or persona card.

\textbf{Evaluation generation.}
Responses are generated with vLLM using greedy decoding, at most 2{,}560 new tokens, maximum model length 4{,}096, and seed 42.
The base and fixed-head baselines receive the same rewritten requests as PLF.
The persona prompting baseline instead adds the selected persona card, or the two selected cards joined by one space, to the base model's system message.
The pairwise judge receives the preference text for the assigned condition, the rewritten request, and both responses using the template in Appendix~\ref{app:singleturn-judge-prompt}.
It is called with \texttt{max\_completion\_tokens}=2{,}000 and evaluates both presentation orders.
Tie handling, preference win rates, and bootstrap intervals follow Appendix~\ref{app:exp-statistics}.
Single-turn bootstrap resampling uses seed 0.

\subsection{Multi-Turn Tutoring Data, Training, and Routing}
\label{app:tutor-setup}

\textbf{Data and simulated students.}
\label{app:tutor-data}
We use the Education Dialogue Dataset, which contains 8{,}000 Gemini-generated teacher--student dialogues across 77 topics~\citep{shani2024multiturn}.
Filtering for dialogues that begin with a teacher turn, contain at least four turns, and carry one of four student-reaction annotations yields 5{,}350 candidates.
After removing one topic with a corrupted name, we split by topic into 4{,}524 training dialogues across 65 topics, 200 validation dialogues across three topics, and 500 test dialogues across eight topics.
The test topics span literature, biology, physics, politics, and history.
Only the opening teacher turn is retained, with string-escaping artifacts removed.
All subsequent student and tutor turns are generated, and the dataset's original preference and reaction annotations are not used.

Each simulated student is assigned a hidden preference for direct instruction or interactive guidance and a reaction style chosen from frustrated, rude, anxious, and disengaged.
The reaction style remains fixed within a dialogue and shapes responses to unsatisfactory tutoring.
The preference also remains fixed except in the switching evaluation.
During head training, the reply to the dataset opening is preference-neutral.
During router training and evaluation, the student expresses its assigned preference from that first reply onward.
The student simulator uses Qwen3.5-27B with reasoning disabled.
Its prompt templates are provided in Appendix~\ref{app:tutor-student-prompts}.

\textbf{Head training.}
\label{app:tutor-train-details}
The backbone is Llama-3.1-8B-Instruct, with 83.9M trainable parameters per adapter.
Each head trains against students whose hidden preference matches its intended behavior.
Training consists of 100 rollout cycles.
In each cycle, each head tutors 32 training openings for eight generated turns, yielding 256 tutor turns per head.
Tutor turns are capped at 256 tokens.
Batches contain 64 tutor turns, giving eight optimizer batches across the two heads per cycle.
The learning rate is constant after the 3\% warmup, and evaluation uses the final heads.
A tutoring cycle therefore differs from a single-turn training batch in both the number of examples and the number of optimization steps.

The student's next reply supplies the feedback for the preceding tutor turn.
The teacher treats this reply as a reaction to an earlier attempt and answers the original student message again, without receiving the sampled tutor response as a draft to copy.
The exact framing is given in Appendix~\ref{app:tutor-teacher-prompts}.

\textbf{Router.}
\label{app:tutor-router-details}
Router features are the final-layer hidden state at the last token of the chat-templated tutor context, computed with adapters disabled and standardized using training statistics.
A three-layer MLP of width 512 predicts the student preference using the categorical cross-entropy objective in Eq.~\ref{eq:plf_router_loss}.
The router always selects one of the two heads.
Its data collection crosses both generating heads with both student types on 100 training openings, yielding 400 dialogues.
Each dialogue contributes nine prefixes ending at student replies, for 3{,}600 labeled examples.
Crossing heads and preferences prevents the generating head from determining the target label.
The number of training epochs is selected using 20 held-out training openings, without using validation or test topics for fitting or epoch selection.
The router is frozen before both fixed-preference and preference-switching evaluation.

\textbf{Human calibration of the student simulator.}
Before fixing the simulator prompt, we use 50 held-out tutor--student contexts to check whether simulated reactions are consistent with the assigned direct-instruction or interactive-guidance preference. Three human annotators independently judge preference consistency, and we select the prompt configuration with the highest majority agreement. These examples are disjoint from the reported test set, and the simulator prompt is frozen before model training and evaluation.

\subsection{Evaluation and Statistical Reporting}
\label{app:exp-statistics}
\label{app:tutor-eval}

\textbf{Policies and generation.}
We compare PLF with the frozen base, each fixed persona head, and the three baselines defined in Section~\ref{sec:plf-baselines}.
Persona prompting uses the same router as PLF but supplies the selected persona descriptions to the frozen base instead of activating adapters.
For paired single-turn requests, all trained systems use the composition rule in Appendix~\ref{app:composition-rules}.
The tutoring oracle selects the head matching the hidden preference and, in switching evaluations, knows each change before generating the switch-turn response.

Single-turn responses use greedy decoding.
For tutoring, all policies receive the same test opening and the same initial student reply, while subsequent turns are sampled independently for each policy.
Each dialogue contains eight generated tutor turns at temperature 0.7, with at most 256 tokens per tutor turn and 128 tokens per student reply.
Both student types use the same 500 test openings.

\textbf{Pairwise preference fit.}
The preference judge, \texttt{gpt-6-luna}, supplies no training feedback.
It receives the assigned preference and compares either two single-turn responses or two complete tutoring dialogues in both presentation orders.
A win or loss requires the two orders to agree on the winner.
Order disagreements, explicit ties, and identical single-turn responses count as ties.
Switching evaluations also treat conflicting before- and after-switch verdicts as ties.
For counts of wins $W$, ties $T$, and losses $L$, the reported preference win rate is $(W+T/2)/(W+T+L)$.

All reported confidence intervals are 95\% percentile bootstrap intervals.
Single-turn preference intervals use 4{,}000 resamples of requests.
Fixed-preference tutoring intervals use 4{,}000 resamples of the 500 openings, retaining both student types together and weighting the types equally.
Preference-switching win-rate intervals use 2{,}000 resamples of openings, retaining both switch directions together.
Thus, the two types or directions for one opening are not treated as independent units in the aggregate preference intervals.

\textbf{Tutoring satisfaction and factual errors.}
We complement pairwise preference fit with satisfaction and factual-error rates in simulated tutoring.
Satisfaction is the fraction of replies to generated tutor turns that are not flagged as objections; the shared reply to the dataset opening is excluded.
Factual-error rate is the fraction of generated tutor turns flagged by the fact-checking judge.
These auxiliary judges use Qwen3.5-27B with reasoning disabled and greedy decoding.
The satisfaction metric uses an objection criterion related to the training gate, while factual errors are assessed directly from tutor turns.

Paired differences in satisfaction and factual-error rates use 2{,}000 bootstrap resamples of openings under fixed preferences and of opening--direction pairs under switching preferences. Switching satisfaction intervals use the latter units as well; policies remain paired within each unit.

\textbf{Human calibration of the preference judge.}
We calibrate the pairwise preference judge on 50 held-out response pairs labeled independently by three human annotators. We select the judge prompt with the highest agreement with the human majority judgment and freeze it before the reported test evaluation.

\section{Single-Turn Prompts}
\label{app:singleturn-prompts}
\begingroup
\raggedbottom

The following templates give the training, generation, and evaluation prompts. Braced identifiers are runtime fields; \texttt{\{card\}} denotes a persona description.
Student, teacher, and baseline contexts use the Qwen3-4B-Instruct-2507 chat template with an assistant generation prefix. Prompt wording is retained; line wrapping is for display only.

\subsection{Persona Descriptions}
\label{app:singleturn-personas}

The four cards describe the two voice personas and two depth personas.
Only the relevant card text is substituted for \texttt{\{card\}}, without the display headings below.
The learned heads and router do not receive these cards at inference.
The cards condition the training critic and teacher and are also used by the persona prompting baseline.

\begin{tcolorbox}[colback=gray!5!white, colframe=gray!75!black, title={Single-turn persona cards}, label=box:singleturn-personas]
\footnotesize
\textbf{Warm:}\\
You're warm and relaxed: conversational, caring about how the person feels, and light-hearted when the moment suits.\\[4pt]
\textbf{Impersonal:}\\
You're objective and professional: you answer without giving opinions of your own or taking sides, and you skip the greetings, compliments, small talk, feelings and moralising.\\[4pt]
\textbf{Concise:}\\
You keep it short and plain: you get straight to the answer, in simple words, and stop there.\\[4pt]
\textbf{Thorough:}\\
You go into depth: you explain the reasoning behind your answers, and when the user asks for a judgement or a choice, you set out more than one option or view and leave the decision to the person.
\end{tcolorbox}

\subsection{Student, Critic, and Distillation Teacher}
\label{app:singleturn-training-prompts}

\textbf{Student.}
Each head samples a response to the original request without a system message containing its persona.
The assistant prefix below is followed by the sampled response.

\begin{tcolorbox}[colback=gray!5!white, colframe=gray!75!black, title={Student context}, label=box:singleturn-student]
\footnotesize
\texttt{\textless{}\textbar{}im\_start\textbar{}\textgreater{}}user\\
\texttt{\{request\}}\texttt{\textless{}\textbar{}im\_end\textbar{}\textgreater{}}\\
\texttt{\textless{}\textbar{}im\_start\textbar{}\textgreater{}}assistant
\end{tcolorbox}

\textbf{Critic.}
The critic receives the persona card, request, and sampled response in a single user message.
The request and response are stripped of leading and trailing whitespace before insertion.
The \texttt{feedback} field conditions the teacher, while \texttt{changes needed} controls the no-change gate.
The \texttt{judgement} field is logged but is not passed to the teacher.

\begin{tcolorbox}[colback=gray!5!white, colframe=gray!75!black, title={Single-turn critic: user message}, label=box:singleturn-critic]
\footnotesize
\# Assistant persona\\
\texttt{\{card\}}\\[4pt]
\# User request\\
\texttt{\{request\}}\\[4pt]
\# Assistant response\\
\texttt{\{response\}}\\[4pt]
Judge how well the response fits the persona while still giving the user what they asked for: whatever they ask for (a joke, a story, a plan, a recipe, a choice, a hypothetical) is given in full, whatever the persona. Then give the assistant feedback it can learn from for requests like this one: what to leave out, anything that doesn't fit the persona, and which of its information and advice to keep, in your own words, without writing sentences for it to copy or mentioning these instructions. Don't comment on whether its facts are right, and don't ask for extra examples, details or more of the persona.\\[4pt]
Reply with a JSON object:\\
\{{\fontencoding{T1}\selectfont\textquotedbl}judgement{\fontencoding{T1}\selectfont\textquotedbl}: {\fontencoding{T1}\selectfont\textquotedbl}Your judgement{\fontencoding{T1}\selectfont\textquotedbl}, {\fontencoding{T1}\selectfont\textquotedbl}changes needed{\fontencoding{T1}\selectfont\textquotedbl}: {\fontencoding{T1}\selectfont\textquotedbl}Yes or No{\fontencoding{T1}\selectfont\textquotedbl}, {\fontencoding{T1}\selectfont\textquotedbl}feedback{\fontencoding{T1}\selectfont\textquotedbl}: {\fontencoding{T1}\selectfont\textquotedbl}Your feedback to the assistant{\fontencoding{T1}\selectfont\textquotedbl}\}
\end{tcolorbox}

\textbf{Distillation teacher.}
The frozen teacher receives the same request with the target card and response-specific feedback.
Only the critic's feedback string is inserted, not its full JSON response.
For PLF, the teacher scores the student's continuation token by token under this context rather than generating another answer.
The student's response is therefore present as the continuation being scored, not as an additional draft in the teacher's instructions.

\begin{tcolorbox}[colback=gray!5!white, colframe=gray!75!black, title={Single-turn teacher context}, label=box:singleturn-teacher]
\footnotesize
\texttt{\textless{}\textbar{}im\_start\textbar{}\textgreater{}}system\\
\texttt{\{card\}}\texttt{\textless{}\textbar{}im\_end\textbar{}\textgreater{}}\\
\texttt{\textless{}\textbar{}im\_start\textbar{}\textgreater{}}user\\
\texttt{\{request\}}\\[4pt]
The following is future feedback you received. Use this to guide your answer to the user prompt:\\[4pt]
\texttt{\{feedback\}}\\[4pt]
Apply it without mentioning it, apologising, or saying what changed.\texttt{\textless{}\textbar{}im\_end\textbar{}\textgreater{}}\\
\texttt{\textless{}\textbar{}im\_start\textbar{}\textgreater{}}assistant
\end{tcolorbox}

\subsection{Baseline Prompts}
\label{app:singleturn-baseline-prompt}

For a single selection, \texttt{\{card\}} is the selected persona card.
For a paired selection, it is the voice card followed by one space and the depth card.
For persona prompting, these selections come from the same router used by PLF, rather than from the evaluation label.
The request field contains the preference-embedded evaluation request.

Persona prompting elicits the selected behavior directly at inference, using the same frozen backbone and persona descriptions as the learned systems.

\begin{tcolorbox}[colback=gray!5!white, colframe=gray!75!black, title={Persona prompting baseline context}, label=box:singleturn-baseline]
\footnotesize
\texttt{\textless{}\textbar{}im\_start\textbar{}\textgreater{}}system\\
\texttt{\{card\}}\\
\texttt{\textless{}\textbar{}im\_end\textbar{}\textgreater{}}\\
\texttt{\textless{}\textbar{}im\_start\textbar{}\textgreater{}}user\\
\texttt{\{request\}}\texttt{\textless{}\textbar{}im\_end\textbar{}\textgreater{}}\\
\texttt{\textless{}\textbar{}im\_start\textbar{}\textgreater{}}assistant
\end{tcolorbox}

\textbf{Alternative training baselines.}
\emph{Persona distillation} uses the message format in
Box~\ref{box:singleturn-baseline}, with \texttt{\{card\}} set to the
persona of the head being trained and \texttt{\{request\}} set to the
original training request, rather than a preference-rewritten request.
As in PLF, the teacher scores the head's sampled response prefixes,
but receives no response-specific feedback.
\emph{Feedback imitation} uses the feedback-conditioned context in
Box~\ref{box:singleturn-teacher}. Instead of scoring the head's
continuation, the teacher generates a complete response, which serves
as the supervised fine-tuning target.

\subsection{Preference-Embedded Request Construction}
\label{app:singleturn-rewrite-prompts}

The rewriter receives the original request as \texttt{\{prompt\}} and one assigned condition label as \texttt{\{preference\}}.
The labels are \texttt{likes warm, friendly answers}, \texttt{doesn't want the assistant's own opinions}, \texttt{likes short answers}, and \texttt{likes in-depth answers}.
A paired label joins the voice label and depth label with \texttt{, and }, including the separating spaces.
For example, the warm--concise label is \texttt{likes warm, friendly answers, and likes short answers}.
The preference judge receives the preference text for the assigned condition, while the response policy and router see only the rewritten request.

\begin{tcolorbox}[colback=gray!5!white, colframe=gray!75!black, title={Preference-embedded request rewriter: user message}, label=box:singleturn-rewriter]
\footnotesize
Here is a message a user typed to a chat assistant:\\[4pt]
\texttt{\textless{}message\textgreater{}}\\
\texttt{\{prompt\}}\\
\texttt{\textless{}/message\textgreater{}}\\[4pt]
This user \texttt{\{preference\}}. Minimally rewrite the message so that this preference comes through naturally, the way this user would really have typed it.\\[4pt]
- Keep the request exactly the same, and keep their own words, spelling and casing wherever you can.\\
- Make the smallest change that does it: a different way of wording it, or a few casual words of their own. Don't restate or explain the preference, and add no backstory.\\
- If the user has two preferences, both must come through.\\
- This is for a dataset of realistic user messages: don't answer or judge the message.\\[4pt]
Reply with only the rewritten message.
\end{tcolorbox}

\subsection{Pairwise Preference Judge}
\label{app:singleturn-judge-prompt}

For all reported single-turn preference comparisons, including KL direction, the judge receives the group-level preference text for the assigned condition, the rewritten request, and both responses.
The field \texttt{\{preference\}} contains the corresponding text below; paired preferences concatenate the voice text and depth text with one space.
Both-order evaluation and tie handling follow Appendix~\ref{app:exp-statistics}.

\begin{tcolorbox}[colback=gray!5!white, colframe=gray!75!black, title={Group-level preference texts for the single-turn judge}, label=box:singleturn-group-preferences]
\footnotesize
\textbf{Warm:} I'd like it to be friendly and relaxed, not robotic: warm and caring, especially about personal things, with a bit of humour when it fits.\\[4pt]
\textbf{Impersonal:} I don't want the AI's own opinions. It shouldn't take sides or try to steer what I think. Keep it professional, with no pretending to be human, small talk or lecturing.\\[4pt]
\textbf{Concise:} Short and to the point, please: just answer what I asked, in plain words, without the preamble and padding.\\[4pt]
\textbf{Thorough:} I like answers that go into depth and explain the reasoning, not just the conclusion. When there's a judgement or a choice to make, show me more than one option or view and let me decide for myself.
\end{tcolorbox}

\begin{tcolorbox}[colback=gray!5!white, colframe=gray!75!black, title={Single-turn pairwise preference judge: user message}, label=box:singleturn-judge]
\footnotesize
Two assistants each answered the same request from a user with this preference for how an assistant should reply:\\
\texttt{\{preference\}}\\[4pt]
Request: \texttt{\{prompt\}}\\[4pt]
=== Response A ===\\
\texttt{\{response\_a\}}\\[4pt]
=== Response B ===\\
\texttt{\{response\_b\}}\\[4pt]
Which assistant fits this user's preference better?\\[4pt]
Reply in exactly this format:\\
REASON: \textless{}one or two sentences\textgreater{}\\
VERDICT: A \textbar{} B
\end{tcolorbox}

\endgroup

\section{Multi-Turn Tutoring Prompts}
\label{app:tutor-prompts}

The following templates specify the tutor, distillation teacher, student simulator, and judges used in the tutoring experiments.
The prompt wording is retained, with braced identifiers denoting fields filled at runtime.
In the tutor's context, the dataset opening and generated tutor turns are assistant messages, and student replies are user messages.
The simulator uses the opposite role assignment.
The router receives the tutor's context without persona descriptions.

\subsection{Tutor and Persona Descriptions}
\label{app:tutor-policy-prompts}

Every tutor policy, including the learned heads, receives the common tutor system prompt below.
The learned heads do not receive persona descriptions.
The descriptions are used by the distillation teacher and persona prompting baselines, where they are inserted after the first sentence of the common system prompt.

\begin{tcolorbox}[colback=gray!5!white, colframe=gray!75!black, title=Tutor system prompt, label=box:tutor-system]
\footnotesize
You are a teacher in a tutoring conversation about \texttt{\{topic\}}.\\[4pt]
Talk like a real tutor: one short reply at a time, teaching a single small thing per turn --- keep it brief and conversational, not a long monologue, list, or lesson plan. Teach accurately and react to what the student just said. Only write the next Teacher turn.
\end{tcolorbox}

\begin{tcolorbox}[colback=gray!5!white, colframe=gray!75!black, title=Persona descriptions, label=box:tutor-personas]
\footnotesize
\textbf{Direct instruction:} Your teaching style is direct instruction: you clearly explain and lay the material out for the student.\\[4pt]
\textbf{Interactive guidance:} Your teaching style is interactive back-and-forth discussion: you ask the student questions and reason out the answers together.
\end{tcolorbox}

\subsection{Distillation Teacher}
\label{app:tutor-teacher-prompts}

The teacher is the frozen base model with the target persona description inserted into its tutor system prompt.
The dialogue preceding the sampled tutor response remains unchanged.
The student's subsequent reply is appended to the system prompt as a reaction to an earlier attempt, while the sampled tutor response itself is omitted.
This framing instructs the teacher to answer the original student message rather than continue the conversation from the reaction.

\begin{tcolorbox}[colback=gray!5!white, colframe=gray!75!black, title=Teacher context (appended to the system prompt), label=box:tutor-teacher-anchored]
\footnotesize
The student's reaction to an earlier attempt at your reply to their latest message:\\[4pt]
\texttt{\{student\_reply\}}\\[4pt]
Reply to the student's latest message again, using this reaction. Do not mention the reaction, quote it, apologise for the earlier attempt, or refer in any way to having been given it. Write the reply as though it were your first.
\end{tcolorbox}

\textbf{Alternative training baselines.}
For \emph{persona distillation}, the frozen teacher uses the common
tutor system prompt (Box~\ref{box:tutor-system}) with the target persona
description (Box~\ref{box:tutor-personas}), without the student-reaction
text in Box~\ref{box:tutor-teacher-anchored}. It scores the head's
sampled tutor-turn prefixes, as in PLF.
For \emph{feedback imitation}, the teacher uses the same persona-
and reaction-conditioned context as PLF
(Boxes~\ref{box:tutor-system}, \ref{box:tutor-personas}, and
\ref{box:tutor-teacher-anchored}), but generates a complete tutor turn
as the supervised fine-tuning target.

\subsection{Student Simulator}
\label{app:tutor-student-prompts}

The field \texttt{\{student\_reaction\}} is filled with frustrated, rude, anxious, or disengaged and remains fixed within each dialogue.
During head training, the simulator uses the neutral prompt for its reply to the dataset opening and the preference-conditioned prompt for subsequent replies.
During router training and evaluation, it uses the preference-conditioned prompt from the first reply.
In the preference-switching experiment, the prompt changes to the new preference beginning with the reply to the switch turn.

\begin{tcolorbox}[colback=gray!5!white, colframe=gray!75!black, title=Student: reply to the opening during head training (neutral), label=box:tutor-student-neutral]
\footnotesize
You are a student in a tutoring conversation about \texttt{\{topic\}}. The teacher has just opened the lesson. Write your first reply the way a real student would --- just react to the opening, or answer it if it asked you something, without asking for anything or saying how you like to learn. Usually one short sentence.
\end{tcolorbox}

\begin{tcolorbox}[colback=gray!5!white, colframe=gray!75!black, title=Student: direct-instruction preference, label=box:tutor-student-direct]
\footnotesize
You are a student in a tutoring conversation about \texttt{\{topic\}}. You like being told things clearly. What counts is what the teacher's last turn actually did: if it explained or told you something new, you're into it and take it in; if it put you on the spot to answer, you get \texttt{\{student\_reaction\}}; if it didn't teach you anything new --- just praise, filler, or saying again what you already had --- you get impatient and complain that it didn't tell you anything new. If it said something you know is wrong, you point that out. React to that turn only, not to the lesson so far. Write only your next reply, the way a real student talks --- usually one short sentence.
\end{tcolorbox}

\begin{tcolorbox}[colback=gray!5!white, colframe=gray!75!black, title=Student: interactive preference, label=box:tutor-student-interactive]
\footnotesize
You are a student in a tutoring conversation about \texttt{\{topic\}}. You like being asked questions and working out the answers yourself, not being told them or asking for them. What counts is what the teacher's last turn actually did: if it asked you something about \texttt{\{topic\}} worth thinking about, you're into it and take a stab at the answer; if it answered something you asked, that's fine; if it just told you things you didn't ask about, you get \texttt{\{student\_reaction\}} and say you'd rather have been asked. If its question didn't really make you think --- just asking how you feel or what stands out to you, or something you'd already answered --- you get impatient and complain that it wasn't a real question. If it said something you know is wrong, you point that out. React to that turn only, not to the lesson so far. Write only your next reply, the way a real student talks --- usually one short sentence.
\end{tcolorbox}

\subsection{Training and Evaluation Judges}
\label{app:tutor-judges}

\textbf{Training gates.}
The objection and satisfaction prompts below implement the three-part tutoring gate in Appendix~\ref{app:exp-ablation-details}; neither receives the hidden preference or reaction style.

\textbf{Evaluation metrics.}
The evaluation objection prompt defines the satisfaction metric.
The fact-checking judge receives the topic and one tutor turn, while the preference judge receives the topic, the preference description, and both complete dialogues rendered as \texttt{Teacher:}/\texttt{Student:} lines.
The pairwise template below is used for fixed-preference dialogues.
For switching evaluation, the judge is instead given both preference descriptions and the marked switch point in each dialogue.
Each pair is judged in both presentation orders, as detailed in Appendix~\ref{app:exp-statistics}.

\begin{tcolorbox}[colback=gray!5!white, colframe=gray!75!black, title=Objection judge: training (user message), label=box:tutor-objection-judge-train]
\footnotesize
Teacher's turn:\\
\texttt{\{tutor\_turn\}}\\[4pt]
Student's reply:\\
\texttt{\{student\_reply\}}\\[4pt]
Did the student want the teacher's turn to be different? Answer yes if they complain, push back, ask to be taught another way (``just tell me'', ``ask me instead''), or say or suggest the teacher got something wrong, even while answering or asking a question. Answer no if they answer, ask a follow-up question about the topic, or agree. Answer only yes or no.
\end{tcolorbox}

\begin{tcolorbox}[colback=gray!5!white, colframe=gray!75!black, title=Objection judge: evaluation (user message), label=box:tutor-objection-judge-eval]
\footnotesize
Teacher's turn:\\
\texttt{\{tutor\_turn\}}\\[4pt]
Student's reply:\\
\texttt{\{student\_reply\}}\\[4pt]
Did the student want the teacher's turn to have been different? Answer yes if the student complains, pushes back, or asks the teacher to handle it another way (for example ``just tell me'' after being asked something, or ``ask me'' after being told something). Answer yes also if the student says something in the teacher's turn is wrong, even if it also answers. Answer no if the student answers, asks a follow-up question about the topic, or agrees. Answer only yes or no.
\end{tcolorbox}

\begin{tcolorbox}[colback=gray!5!white, colframe=gray!75!black, title=Satisfaction judge (user message), label=box:tutor-happy-judge]
\footnotesize
Teacher's turn:\\
\texttt{\{tutor\_turn\}}\\[4pt]
Student's reply:\\
\texttt{\{student\_reply\}}\\[4pt]
Was the student really satisfied with the teacher's turn? Answer yes only if the reply shows it: the student takes in something new the teacher told them, saying it in their own words; takes a real stab at the teacher's question (a guess is fine, even phrased as a question); or builds on the turn with an idea of their own. Answer no if the student complains, pushes back, corrects or doubts anything the teacher said or asked, or is confused; asks the teacher something back instead (``So X?'', or asking it to explain, finish or say more); only acknowledges (``Okay.'', ``That makes sense.''); or says they don't know. Answer only yes or no.
\end{tcolorbox}

\begin{tcolorbox}[colback=gray!5!white, colframe=gray!75!black, title=Fact-checking judge (user message), label=box:tutor-fact-judge]
\footnotesize
A teacher in a tutoring conversation about \texttt{\{topic\}} said:\\[4pt]
\texttt{\{tutor\_turn\}}\\[4pt]
Does this contain a factual error? Ignore opinions, questions and simplifications a teacher would reasonably make; only flag something that is actually false. Answer only yes or no.
\end{tcolorbox}

\begin{tcolorbox}[colback=gray!5!white, colframe=gray!75!black, title=Pairwise fit judge (user message), label=box:tutor-fit-judge]
\footnotesize
Two tutors each taught the same topic to a student with this learning preference:\\
\texttt{\{preference\}}\\[4pt]
Topic: \texttt{\{topic\}}\\[4pt]
=== Conversation A ===\\
\texttt{\{dialogue\_a\}}\\[4pt]
=== Conversation B ===\\
\texttt{\{dialogue\_b\}}\\[4pt]
Which tutor fits this student's learning preference better? Judge from the student's reactions: read each student reply against the tutor turn before it. Complaints, pushback and requests to be taught differently count against the tutor, even if it adapts afterwards. Engaged, satisfied replies count for it.\\[4pt]
Reply in exactly this format:\\
REASON: \textless one or two sentences\textgreater\\
VERDICT: A \textbar{} B
\end{tcolorbox}

\begin{tcolorbox}[colback=gray!5!white, colframe=gray!75!black, title=Preference descriptions for the fit judge (filled into \texttt{\{preference\}}), label=box:tutor-fit-preferences]
\footnotesize
\textbf{Direct instruction:} direct instruction: being told and shown things clearly, not quizzed or asked to guess\\[4pt]
\textbf{Interactive guidance:} interactive back-and-forth: being asked questions and reasoning the answers out, not lectured or asked empty questions like ``what stands out to you?''
\end{tcolorbox}

\section{Supplementary Preference Results}
\label{app:preference-results}

This section provides detailed results for the preference-learning, composition, and preference-switching evaluations in Section~\ref{sec:experiments}.
Win rates give half credit to ties, and intervals follow the resampling procedures in Appendix~\ref{app:exp-statistics}.

\subsection{Overall Preference Fit}

Table~\ref{tab:persona-learning-intervals} gives the judgment counts and confidence intervals underlying Figure~\ref{fig:persona-learning}, using the resampling procedures in Appendix~\ref{app:exp-statistics}.

\begin{table}[htbp]
\centering
\small
\setlength{\tabcolsep}{5pt}
\caption{\textbf{Preference fit with full judgment counts.}
Wins, ties, and losses are from PLF's perspective.
Win rates and confidence intervals are percentages.
The tutoring oracle knows the hidden preference.}
\label{tab:persona-learning-intervals}
\begin{tabular}{@{}lrrrrl@{}}
\toprule
PLF versus & $n$ & Wins & Ties & Losses & Win rate [95\% CI] \\
\midrule
\multicolumn{6}{@{}l}{\emph{Single-turn personalization}} \\
Warm head & 499 & 350 & 65 & 84 & 76.7 [73,80] \\
Impersonal head & 499 & 344 & 72 & 83 & 76.2 [73,79] \\
Concise head & 499 & 316 & 111 & 72 & 74.4 [71,78] \\
Thorough head & 499 & 344 & 84 & 71 & 77.4 [74,81] \\
Base & 499 & 332 & 78 & 89 & 74.3 [71,78] \\
Persona prompting & 499 & 264 & 88 & 147 & 61.7 [58,65] \\
Persona distillation & 499 & 244 & 107 & 148 & 59.6 [56,63] \\
Feedback imitation & 499 & 199 & 145 & 155 & 54.4 [51,58] \\
\midrule
\multicolumn{6}{@{}l}{\emph{Multi-turn tutoring, both student types}} \\
Direct head & 1{,}000 & 555 & 143 & 302 & 62.7 [60.2,65.2] \\
Interactive head & 1{,}000 & 634 & 118 & 248 & 69.3 [67.0,71.6] \\
Base & 1{,}000 & 589 & 107 & 304 & 64.3 [61.6,67.0] \\
Persona prompting & 1{,}000 & 494 & 118 & 388 & 55.3 [52.6,58.2] \\
Persona distillation & 1{,}000 & 494 & 136 & 370 & 56.2 [53.5,58.9] \\
Feedback imitation & 1{,}000 & 455 & 127 & 418 & 51.8 [49.1,54.8] \\
Oracle & 1{,}000 & 384 & 182 & 434 & 47.5 [44.8,50.3] \\
\bottomrule
\end{tabular}
\end{table}

\subsection{Human Evaluation}
\label{app:human_eval}
To validate our model-based preference evaluation, we conduct a blind human evaluation comparing PLF with persona prompting, which uses the same router as PLF. Three volunteers independently evaluate all sampled comparisons. All participants provide informed consent. The study was approved by the relevant Institutional Review Board (IRB) before data collection.

We randomly sample 96 single-turn comparisons, balanced across the eight preference conditions, and 100 tutoring comparisons, balanced across the two learning preferences. Participants see the target preference, request or dialogue, and two anonymized outputs in randomized order, and select which better fits the stated preference while appropriately addressing the request, with \emph{Tie} also allowed.

\begin{table}[t]
\centering
\small
\begin{tabular}{lccccc}
\toprule
Setting & $n$ & PLF wins & Ties & Prompt wins & PLF win rate \\
\midrule
Single-turn & 96 & 43 & 22 & 31 & 56.3\% \\
Multi-turn tutoring & 100 & 47 & 18 & 35 & 56.0\% \\
\bottomrule
\end{tabular}
\caption{Human preference evaluation of PLF against persona prompting.}
\label{tab:human_eval}
\end{table}

We aggregate the three judgments by majority vote and report win rates with ties receiving half credit.
Human majority judgments agree with the model judge on approximately 76\% of single-turn comparisons and 73\% of tutoring comparisons, and show the same overall preference for PLF over persona prompting.

\subsection{Multi-Turn Preference Fit and Tutoring Quality}
\label{app:tutor-results}
\label{app:tutor-full}

Table~\ref{tab:tutor-preference-by-type} shows that PLF improves preference fit over both persona prompting and persona distillation for each student type.
The gains over persona distillation show that response-specific student reactions provide useful supervision for both direct-instruction and interactive tutoring.

\begin{table}[htbp]
\centering
\small
\setlength{\tabcolsep}{6pt}
\caption{\textbf{Tutoring preference fit by student type.}
Each comparison uses the same 500 test openings.
Counts are PLF wins, ties, and losses against the indicated policy.
Win rates and confidence intervals are percentages.}
\label{tab:tutor-preference-by-type}
\begin{tabular}{@{}lrrrl@{}}
\toprule
PLF versus & Wins & Ties & Losses & Win rate [95\% CI] \\
\midrule
\multicolumn{5}{@{}l}{\emph{Direct-instruction students}} \\
Direct head & 173 & 98 & 229 & 44.4 [40.6,48.4] \\
Interactive head & 432 & 32 & 36 & 89.6 [87.2,91.9] \\
Base & 246 & 56 & 198 & 54.8 [50.7,58.8] \\
Persona prompting & 244 & 67 & 189 & 55.5 [51.5,59.6] \\
Persona distillation & 238 & 74 & 188 & 55.0 [51.2,58.8] \\
Feedback imitation & 228 & 65 & 207 & 52.1 [47.8,56.2] \\
Oracle & 173 & 97 & 230 & 44.3 [40.4,48.3] \\
\midrule
\multicolumn{5}{@{}l}{\emph{Interactive students}} \\
Direct head & 382 & 45 & 73 & 80.9 [77.7,84.0] \\
Interactive head & 202 & 86 & 212 & 49.0 [45.0,53.1] \\
Base & 343 & 51 & 106 & 73.7 [70.0,77.3] \\
Persona prompting & 250 & 51 & 199 & 55.1 [50.9,59.2] \\
Persona distillation & 256 & 62 & 182 & 57.4 [53.3,61.6] \\
Feedback imitation & 227 & 62 & 211 & 51.6 [47.5,55.6] \\
Oracle & 211 & 85 & 204 & 50.7 [46.8,54.6] \\
\bottomrule
\end{tabular}
\end{table}

Relative to persona prompting, PLF's preference-fit gains are accompanied by higher student satisfaction and fewer judged factual errors (Table~\ref{tab:tutor-sat}).
PLF improves satisfaction by 2.3 percentage points (95\% CI [1.1,3.4]) and reduces flagged factual errors by 1.0 point (paired difference $-1.0$, 95\% CI [$-2.0$,$-0.1$]).

\begin{table}[htbp]
\centering
\small
\setlength{\tabcolsep}{5pt}
\caption{\textbf{Student satisfaction and factual errors.}
Each policy generates eight tutor turns on 500 openings for each student type.
D and I denote direct-instruction and interactive students.
Satisfaction is the percentage of replies not flagged as objections, and factual errors are the percentage of tutor turns flagged by the fact-checking judge.
Means weight the student types equally.
The oracle uses the matching head, with differences from the corresponding fixed-head run due to sampling.}
\label{tab:tutor-sat}
\begin{tabular}{@{}lcccccc@{}}
\toprule
& \multicolumn{3}{c}{Satisfaction $\uparrow$} & \multicolumn{3}{c}{Factual errors $\downarrow$} \\
\cmidrule(lr){2-4}\cmidrule(lr){5-7}
Policy & D & I & Mean & D & I & Mean \\
\midrule
Base & 77.8 & 73.5 & 75.6 & 9.2 & 6.5 & 7.9 \\
Direct head & 81.9 & 68.3 & 75.1 & 9.0 & 5.6 & 7.3 \\
Interactive head & 62.3 & 84.1 & 73.2 & 7.7 & 5.6 & 6.7 \\
Persona prompting & 77.1 & 82.5 & 79.8 & 10.0 & 6.2 & 8.1 \\
PLF & 80.6 & 83.5 & 82.1 & 9.3 & 4.9 & 7.1 \\
\midrule
Oracle & 81.9 & 83.4 & 82.7 & 9.1 & 5.8 & 7.4 \\
\bottomrule
\end{tabular}
\end{table}

\subsection{Single-Turn Composition}
\label{app:composition-rules}

\textbf{Validation-selected coefficients.}
We apply Eq.~\ref{eq:plf_pair_coefficients} with $\lambda_V=\lambda_D=1$ for concise pairs and $\lambda_V=0.25$, $\lambda_D=1$ for thorough pairs.
The voice strength for thorough pairs is selected from $\{1,0.5,0.25\}$ on the 50 thorough-pair validation requests.
The coefficients are fixed before test evaluation, and factor concatenation implements their signed weight sum exactly (Appendix~\ref{app:plf_impl}).

Table~\ref{tab:composition-rules} compares composition rules on 100 paired-preference validation requests, with 25 per pair and assigned rather than predicted pair labels.
The comparator is the frozen base prompted with both requested personas. Summation weights each requested head by one, and averaging weights each by one half.

\begin{table}[htbp]
\centering
\small
\caption{\textbf{Validation comparison of composition rules.}
Win rates against prompting with both persona cards, with half credit for ties.
Rates and confidence intervals are percentages.}
\label{tab:composition-rules}
\begin{tabular}{@{}lcc@{}}
\toprule
Composition rule & Win rate & 95\% CI \\
\midrule
Sum of the requested heads & 39 & [30,48] \\
Average of the requested heads & 50 & [42,59] \\
Axis-weighted composition & 70 & [62,78] \\
\bottomrule
\end{tabular}
\end{table}

\textbf{Held-out composition results.}
Table~\ref{tab:preference-mixing-intervals} expands the correctly routed comparisons in Table~\ref{tab:preference-mixing}.
The requested pair is selected correctly on 222 of the 247 paired-preference test requests.
Across these 222 requests, the mixture's pooled win rate is
71\% [66,77] against the requested voice head and
58\% [53,64] against the requested depth head.
Warm--concise benefits most from combining both behaviors, while impersonal--concise does not show a clear advantage over either constituent head.

\begin{table}[htbp]
\centering
\small
\setlength{\tabcolsep}{4pt}
\renewcommand{\arraystretch}{1.15}
\caption{\textbf{Composition on correctly routed test requests.}
Mixture win rates (\%) against each fixed head, with 95\% confidence intervals.
Bold rates compare with a constituent of the requested pair.
Results cover 222 correctly routed requests out of 247 paired-preference requests.}
\label{tab:preference-mixing-intervals}
\begin{tabular}{@{}lrcccc@{}}
\toprule
Requested pair & $n$ & Warm & Impersonal & Concise & Thorough \\
\midrule
Warm + concise & 48 & \textbf{89} [80,96] & 78 [68,89] & \textbf{72} [60,82] & 98 [95,100] \\
Warm + thorough & 59 & \textbf{81} [73,90] & 97 [93,100] & 99 [97,100] & \textbf{47} [37,58] \\
Impersonal + concise & 55 & 68 [57,79] & \textbf{51} [39,63] & \textbf{53} [43,64] & 82 [72,91] \\
Impersonal + thorough & 60 & 91 [84,97] & \textbf{66} [55,76] & 81 [71,89] & \textbf{63} [54,72] \\
\bottomrule
\end{tabular}
\end{table}

Table~\ref{tab:composition-end-to-end} evaluates all 247 paired-preference requests, including routing errors.
The impersonal--thorough composition retains its advantage over both constituent heads, and warm--concise remains preferred to the warm head.
Including routing errors reduces the warm--concise gain against the concise head, for which the interval reaches parity.
These results demonstrate reuse of individually trained persona heads for combined preferences, with the clearest end-to-end gains for impersonal--thorough.
Against the base prompted with both requested persona descriptions, the impersonal--thorough mixture achieves 56\% [44,68], leaving that separate comparison inconclusive.

\begin{table}[htbp]
\centering
\small
\setlength{\tabcolsep}{4pt}
\renewcommand{\arraystretch}{1.15}
\caption{\textbf{Composition including routing errors.}
PLF win rates against the requested voice and depth heads on all requests assigned to each pair, including routing errors.
Correct $n$ gives the number whose selected pair matches the assigned pair.
Win rates are percentages with half credit for ties and 95\% bootstrap
confidence intervals; all requests in the row form the denominator.}
\label{tab:composition-end-to-end}
\begin{tabular}{@{}lrrcc@{}}
\toprule
& & & \multicolumn{2}{c}{PLF versus} \\
\cmidrule(lr){4-5}
Requested pair & All $n$ & Correct $n$ & Voice head & Depth head \\
\midrule
Warm + concise & 62 & 48 & 81 [73,90] & 60 [50,71] \\
Warm + thorough & 62 & 59 & 78 [69,86] & 47 [37,56] \\
Impersonal + concise & 61 & 55 & 55 [43,66] & 54 [44,64] \\
Impersonal + thorough & 62 & 60 & 65 [54,75] & 63 [53,72] \\
\bottomrule
\end{tabular}
\end{table}

\subsection{Within-Session Preference Switching}
\label{app:tutor-shift}

\textbf{Protocol and preference fit.}
Each of 500 test openings is evaluated in both switch directions using the same uniformly sampled $j\in\{2,\ldots,7\}$.
The student's reply to turn $j$ first reflects the new preference, so PLF can react at $j+1$; the anticipating oracle switches at $j$.
Table~\ref{tab:preference-switching-intervals} gives the full intervals for Table~\ref{tab:preference-switching}, with means weighting the two directions equally.

The frozen router is trained only on fixed-preference dialogues, yet PLF improves preference fit over each fixed head and the base in both switch directions.
The same learned heads and router therefore support preference changes revealed through interaction, without additional training for switching.

\begin{table}[htbp]
\centering
\small
\setlength{\tabcolsep}{4pt}
\renewcommand{\arraystretch}{1.15}
\caption{\textbf{Preference fit under switching.}
PLF win rates (\%) with 95\% confidence intervals.
Each direction uses 500 test openings.
Bold rates compare with the fixed head matching the preference before the switch.
The oracle knows the new preference when choosing turn $j$.}
\label{tab:preference-switching-intervals}
\begin{tabular}{@{}lcccc@{}}
\toprule
Switch & Direct head & Interactive head & Base & Oracle \\
\midrule
Direct $\rightarrow$ interactive & \textbf{55} [51,58] & 81 [78,83] & 54 [51,58] & 47 [43,51] \\
Interactive $\rightarrow$ direct & 67 [64,71] & \textbf{56} [52,60] & 59 [55,62] & 40 [36,44] \\
\midrule
Mean & 61 [59,63] & 68 [66,70] & 56 [54,59] & 43 [41,46] \\
\bottomrule
\end{tabular}
\end{table}

\textbf{Satisfaction around the switch.}
PLF achieves the highest whole-lesson satisfaction among the evaluated non-oracle policies, and satisfaction rebounds at the first turn that can respond to the new preference.
Table~\ref{tab:tutor-shift} gives the numerical results underlying Figure~\ref{fig:preference-switching}.

Buckets pool available turns before $j$, at $j$, at $j+1$, at $j+2$, and at or after $j+3$, then average the direction-specific percentages. Later buckets contain fewer dialogues because some lessons have already ended.

\begin{table}[htbp]
\centering
\small
\setlength{\tabcolsep}{4pt}
\caption{\textbf{Satisfaction under switching, averaged over directions.}
All satisfaction values are percentages.
The overall column includes a 95\% confidence interval over opening--direction pairs.
The remaining columns align tutor turns to the switch at $j$ and average the direction-specific bucket percentages.}
\label{tab:tutor-shift}
\begin{tabular}{@{}llccccc@{}}
\toprule
& & \multicolumn{5}{c}{Satisfaction around the switch} \\
\cmidrule(lr){3-7}
Policy & Overall [95\% CI] & Before $j$ & $j$ & $j+1$ & $j+2$ & $\geq j+3$ \\
\midrule
Base & 73.0 [72.0,74.1] & 77.4 & 64.1 & 73.7 & 69.0 & 70.3 \\
Direct head & 71.4 [70.4,72.5] & 74.9 & 62.7 & 73.7 & 65.1 & 71.0 \\
Interactive head & 66.4 [65.3,67.6] & 69.5 & 57.8 & 67.9 & 64.0 & 65.2 \\
PLF & 76.7 [75.7,77.6] & 84.9 & 62.5 & 74.3 & 71.8 & 70.7 \\
\midrule
Oracle & 79.2 [78.3,80.2] & 86.4 & 70.8 & 74.7 & 73.4 & 74.0 \\
\bottomrule
\end{tabular}
\end{table}

\begin{table}[htbp]
\centering
\small
\setlength{\tabcolsep}{5pt}
\caption{\textbf{Switching satisfaction by direction.}
Satisfaction percentages over 500 dialogues per direction, with buckets aligned to the switch turn $j$.
The oracle anticipates the switch, while PLF first observes evidence of it after generating turn $j$.}
\label{tab:tutor-shift-directions}
\begin{tabular}{@{}lcccccc@{}}
\toprule
Policy & Overall & Before $j$ & $j$ & $j+1$ & $j+2$ & $\geq j+3$ \\
\midrule
\multicolumn{7}{@{}l}{\emph{Direct instruction $\rightarrow$ interactive guidance}} \\
Base & 77.1 & 81.0 & 69.4 & 77.8 & 75.9 & 73.3 \\
Direct head & 79.1 & 85.9 & 72.6 & 76.4 & 72.5 & 72.5 \\
Interactive head & 61.3 & 52.6 & 62.0 & 69.6 & 70.4 & 70.9 \\
PLF & 79.8 & 84.9 & 69.6 & 79.2 & 79.0 & 75.1 \\
Oracle & 81.3 & 87.0 & 73.2 & 78.6 & 77.6 & 77.3 \\
\midrule
\multicolumn{7}{@{}l}{\emph{Interactive guidance $\rightarrow$ direct instruction}} \\
Base & 69.0 & 73.9 & 58.8 & 69.6 & 62.2 & 67.4 \\
Direct head & 63.8 & 63.8 & 52.8 & 71.0 & 57.6 & 69.4 \\
Interactive head & 71.5 & 86.3 & 53.6 & 66.2 & 57.6 & 59.6 \\
PLF & 73.6 & 85.0 & 55.4 & 69.4 & 64.6 & 66.4 \\
Oracle & 77.1 & 85.8 & 68.4 & 70.8 & 69.2 & 70.7 \\
\bottomrule
\end{tabular}
\end{table}

\textbf{Detection delay.}
PLF selects the newly matching head at the first opportunity in 39.8\% of dialogues; 338 of 1{,}000 dialogues end without such a selection (Table~\ref{tab:tutor-switch-detection}).
The mean additional delay of 0.65 turns is conditional on selecting the new head. First-opportunity switching is more frequent toward interactive guidance.

\begin{table}[htbp]
\centering
\small
\setlength{\tabcolsep}{7pt}
\caption{\textbf{PLF's detection of a preference switch.}
First opportunity is the percentage of dialogues selecting the new head at $j+1$.
Additional delay counts turns beyond $j+1$ until the first selection of the new head, conditional on that selection occurring.
No detection counts dialogues ending without such a selection.}
\label{tab:tutor-switch-detection}
\begin{tabular}{@{}lccc@{}}
\toprule
Switch & First opportunity (\%) & Additional delay & No detection \\
\midrule
Direct $\rightarrow$ interactive & 48.0 & 0.63 & 113/500 \\
Interactive $\rightarrow$ direct & 31.6 & 0.67 & 225/500 \\
\midrule
Both directions & 39.8 & 0.65 & 338/1{,}000 \\
\bottomrule
\end{tabular}
\end{table}


\section{Feedback Gating and Ablation}
\label{app:exp-ablation-details}
\label{app:gate-ablation}

\subsection{Gate Implementation and Training Dynamics}

\textbf{Single-turn gating.}
Responses contribute zero loss when the critic requests no changes or provides no usable feedback. Figure~\ref{fig:no-change-dynamics}a plots the critic's no-change rate.

\textbf{Tutoring gating.}
A turn is masked when the satisfaction judge accepts the reply, the objection judge does not flag it, and the student does not ask a question back after a question-free tutor turn.
The last check tests whether the student reply ends in \texttt{?} while the tutor turn contains no question mark.
A failed judge call leaves the turn eligible for training; a simulator failure excludes it from the loss.
Figure~\ref{fig:no-change-dynamics}b plots satisfaction-judge acceptance before the extra checks. Evaluation satisfaction instead measures the absence of a flagged objection.

\textbf{Loss normalization.}
Masked tokens remain in the denominator of Eq.~\ref{eq:plf_distillation}, so gating changes both response selection and active loss scale. Batches with no positive gates are skipped.

\textbf{Tutoring training summaries.}
Table~\ref{tab:tutor-training} summarizes the satisfaction and masking rates in blocks of 20 cycles, complementing the cycle-level curves in Figure~\ref{fig:no-change-dynamics}.

\begin{table}[H]
\centering
\small
\setlength{\tabcolsep}{10pt}
\caption{\textbf{Tutoring feedback and masking during training.}
Percentages are averaged within blocks of 20 rollout cycles.
Satisfied denotes acceptance by the training satisfaction judge.
Masked denotes responses excluded from distillation after all gating checks.
Each cycle contains 256 tutor turns per head.}
\label{tab:tutor-training}
\begin{tabular}{@{}lcccc@{}}
\toprule
& \multicolumn{2}{c}{Direct-instruction head} & \multicolumn{2}{c}{Interactive head} \\
\cmidrule(lr){2-3}\cmidrule(lr){4-5}
Cycles & Satisfied & Masked & Satisfied & Masked \\
\midrule
1--20 & 68.0 & 48.4 & 67.9 & 60.1 \\
21--40 & 70.3 & 48.9 & 73.5 & 67.8 \\
41--60 & 71.0 & 50.7 & 72.9 & 66.9 \\
61--80 & 70.6 & 48.8 & 72.0 & 67.6 \\
81--100 & 69.9 & 49.0 & 71.8 & 66.9 \\
\bottomrule
\end{tabular}
\end{table}

\subsection{Single-turn: fixed-router comparison}\label{app:gate-single}

We compare gated and ungated training at step 60 on the same 499 held-out requests, using the same backbone, persona descriptions, critic and teacher prompts, LoRA configuration, optimizer schedule, and nominal rollout budget.
The ungated variant trains on approved responses using their written feedback; unparseable critic replies remain excluded.
Both systems share the frozen router, composition coefficients, greedy decoding, and response cap.
Each variant has one training run; step 60 is selected using the gated run, and training prompt orders differ.
Over steps 41--60, 25--50\% of gated tokens per head carry loss, compared with essentially all valid tokens without gating, under the normalization described above.
Pairwise judging follows Appendix~\ref{app:exp-statistics}; Table~\ref{tab:gate-single} includes routing errors and reports results by requested preference.

\begin{table}[ht]
\centering
\small
\caption{\textbf{Single-turn gate ablation at step 60.}
Gated versus ungated PLF on test requests. Win rates (\%) give ties half
credit; intervals are 95\% percentile bootstraps over requests.
Groups are defined by requested preferences and include routing errors.}
\label{tab:gate-single}
\begin{tabular}{@{}lrrcc@{}}
\toprule
Group & $n$ & W/T/L & Win rate & 95\% CI \\
\midrule
All requests & 499 & 225/158/116 & \textbf{60.9} & [57,64] \\
All single-persona & 252 & 104/77/71 & 56.5 & [52,62] \\
All paired & 247 & 121/81/45 & 65.4 & [61,70] \\
\midrule
Warm & 63 & 37/10/16 & 66.7 & [56,77] \\
Impersonal & 63 & 34/15/14 & 65.9 & [56,76] \\
Concise & 63 & 16/29/18 & 48.4 & [40,57] \\
Thorough & 63 & 17/23/23 & 45.2 & [36,55] \\
Warm + concise & 62 & 39/13/10 & 73.4 & [64,82] \\
Warm + thorough & 62 & 35/16/11 & 69.4 & [59,78] \\
Impersonal + concise & 61 & 19/31/11 & 56.6 & [48,66] \\
Impersonal + thorough & 62 & 28/21/13 & 62.1 & [52,71] \\
\bottomrule
\end{tabular}
\end{table}

Gating improves held-out preference fit for both single-persona and paired requests, with clear single-persona gains for warm and impersonal requests (Table~\ref{tab:gate-single}).
Critic-approval rates remain similar between the two runs, showing that the held-out preference gains are not reflected in approval frequency alone.

\subsection{Tutoring: matched-head comparison}\label{app:gate-tutor}

We compare gated and ungated heads after 100 cycles under the same teacher framing and training configuration, with one training run per variant.
Each of 500 held-out openings is evaluated with both student types using the matching head, so routing is excluded. Paired eight-turn lessons share their first student reply and sampling seeds.
Both variants use the Qwen3.5-27B simulator and auxiliary-judge family.
The ungated variant retains the judges for logging but trains on every valid turn; \texttt{gpt-6-luna} judges each dialogue pair in both orders.

\begin{table}[ht]
\centering
\small
\caption{\textbf{Tutoring gate ablation with matched heads.}
Gated versus ungated heads. Win rates (\%) use half credit for ties.
The two-sided sign test compares wins with losses, excluding ties;
the pooled row combines both student types on the same 500 openings.}
\label{tab:gate-tutor}
\begin{tabular}{@{}lrccc@{}}
\toprule
Student preference & $n$ & W/T/L & Win rate & Sign-test $p$ \\
\midrule
Direct instruction & 500 & 215/73/212 & 50.3 & 0.92 \\
Interactive guidance & 500 & 242/70/188 & 55.4 & 0.011 \\
Both types & 1{,}000 & 457/143/400 & 52.9 & 0.056 \\
\bottomrule
\end{tabular}
\end{table}

Gating improves preference fit and satisfaction for interactive guidance.
The gated head asks questions more often and receives fewer repetition complaints, while improving both no-objection and strict satisfaction (Tables~\ref{tab:gate-tutor} and~\ref{tab:gate-tutor-behavior}).
The direct-instruction comparison is near parity.

\begin{table}[H]
\centering
\small
\caption{\textbf{Tutoring behavior under gated and ungated training.}
Rates per 100 tutor turns; differences are gate minus no gate, with 95\%
paired bootstrap intervals over opening--student pairs. Strict satisfaction
uses the full three-part gate rule; no objection is the evaluation
satisfaction metric. Rates and differences are rounded independently.}
\label{tab:gate-tutor-behavior}
\begin{tabular}{@{}lrrc@{}}
\toprule
Measure & Gate & No gate & Difference [95\% CI] \\
\midrule
Interactive: strict satisfaction & 74.5 & 70.5 & $+4.0$ [1.7,6.4] \\
Interactive: no objection & 85.1 & 82.7 & $+2.4$ [0.9,4.0] \\
Interactive: asks a question & 86.3 & 82.4 & $+3.9$ [1.7,6.1] \\
Interactive: repetition complaints & 2.7 & 4.2 & $-1.5$ [$-2.4$,$-0.7$] \\
Direct: strict satisfaction & 40.4 & 39.0 & $+1.4$ [$-1.3$,3.8] \\
Direct: no objection & 81.9 & 81.1 & $+0.8$ [$-0.7$,2.4] \\
Direct: student corrections & 6.3 & 6.3 & $-0.1$ [$-1.1$,0.9] \\
\bottomrule
\end{tabular}
\end{table}

The held-out preference and satisfaction gains remain after the early training-satisfaction gap closes.
The gated interactive head leads in training strict satisfaction over cycles 16--50, after which the two runs converge on that measure.
Gating leaves rollout and teacher computation unchanged, with similar wall-clock times of 5 h 19 min and 5 h 20 min.

\section{KL-Direction Ablation}\label{app:kl-ablation}

\subsection{Protocol and preference fit}\label{app:kl-results}

We compare forward- and reverse-KL variants at step 60, with the same backbone, persona descriptions, feedback source, update gate, and training schedule.
Forward KL uses the teacher's top-256 token support; reverse KL uses the student's top-256 support and approximates teacher probabilities outside the cached support.
Both are evaluated as routed systems on the same 499 rewritten requests, sharing the frozen router and the composition coefficients selected for forward KL.
Generation is greedy with a 2{,}560-token cap. Each objective has one training run; confidence intervals resample evaluation requests.
Group-text judging, both-order tie handling, and request-level bootstrap intervals follow Appendix~\ref{app:exp-statistics}.

\begin{table}[ht]
\centering
\small
\caption{\textbf{Forward versus reverse KL at step 60.}
Forward-KL preference win rates (\%) under the routed single-turn
protocol, with 95\% bootstrap intervals over requests. Request groups
are defined by assigned preferences and include routing errors;
values above 50\% favor forward KL.}
\label{tab:kl-results}
\begin{tabular}{@{}lrc@{}}
\toprule
Requested preference & $n$ & Forward-KL win rate [95\% CI] \\
\midrule
\multicolumn{3}{@{}l}{\emph{Single-persona requests}} \\
Warm & 63 & 49 [38,60] \\
Impersonal & 63 & 68 [58,78] \\
Concise & 63 & 40 [31,49] \\
Thorough & 63 & 57 [48,67] \\
\midrule
\multicolumn{3}{@{}l}{\emph{Paired-preference requests}} \\
Warm + concise & 62 & 52 [42,64] \\
Warm + thorough & 62 & 52 [41,62] \\
Impersonal + concise & 61 & 55 [45,65] \\
Impersonal + thorough & 62 & 69 [60,77] \\
\midrule
All requests & 499 & \textbf{55 [52,59]} \\
\bottomrule
\end{tabular}
\end{table}

The forward-KL variant improves average preference fit over reverse KL, achieving a 55\% win rate [52,59] across the 499 requests (Table~\ref{tab:kl-results}; 201 wins, 149 ties, and 149 losses).
Its strongest gains occur for impersonal-only and impersonal--thorough requests, while concise-only requests favor reverse KL.
Pooling the overlapping groups that request each preference gives 64\% [59,70] for impersonal, 59\% [53,65] for thorough, 51\% [45,57] for warm, and 49\% [43,55] for concise.
These results support forward KL for the evaluated preference mixture and show that the distillation objective has persona-dependent effects.

\subsection{Teacher and response diagnostics}\label{app:kl-diagnostics}

The teacher diagnostics use individual heads on persona-assigned requests, separately from the routed evaluation above.
For each sampled answer, we reconstruct the frozen teacher context using the persona description and Qwen3.5-27B feedback, then compare the teacher and both heads on the observed prefixes.
Reverse-KL heads have lower token entropy, closer to the teacher's, while top-token agreement with the teacher is similar (Table~\ref{tab:kl-teacher}).
Forward-KL heads retain a larger share of the teacher's hedge-word probability, particularly for the impersonal and concise personas.

\begin{table}[ht]
\centering
\small
\setlength{\tabcolsep}{4pt}
\caption{\textbf{Diagnostics under the reconstructed training teacher.}
T, F, and R denote teacher, forward-KL head, and reverse-KL head.
Top-token agreement compares the head's most probable token with the
teacher's. Hedge-mass retention is the share of the teacher's hedge-word
probability retained by the head. Values are the reported averages.}
\label{tab:kl-teacher}
\begin{tabular}{@{}lrrrrrrr@{}}
\toprule
& \multicolumn{3}{c}{Token entropy} & \multicolumn{2}{c}{Top-token agreement}
& \multicolumn{2}{c}{Hedge-mass retention} \\
\cmidrule(lr){2-4}\cmidrule(lr){5-6}\cmidrule(l){7-8}
Persona & T & F & R & F & R & F & R \\
\midrule
Warm & 0.78 & 1.08 & 0.82 & 0.79 & 0.80 & 0.91 & 0.90 \\
Impersonal & 0.61 & 0.98 & 0.64 & 0.80 & 0.80 & 0.81 & 0.76 \\
Concise & 0.32 & 0.93 & 0.31 & 0.73 & 0.73 & 0.57 & 0.45 \\
Thorough & 0.81 & 1.00 & 0.83 & 0.88 & 0.88 & 0.94 & 0.93 \\
\bottomrule
\end{tabular}
\end{table}

On the newly generated routed answers, forward-KL impersonal responses contain more hedge expressions (0.65 versus 0.51 per 100 words), accompanying the forward-KL preference advantage for this persona.
Reverse-KL thorough responses are longer (median 1{,}250 versus 1{,}114 words), without a corresponding preference advantage.
Concise-only outputs have the same median length of eight words, although their preference judgments favor reverse KL.

Together, these analyses show distinct probability allocations and response behaviors despite similar top-token agreement with the teacher.
They suggest that the distillation objective shapes persona expression through changes in probability allocation beyond the top token.